\documentclass{article}

\usepackage[utf8]{inputenc}
\usepackage[T1]{fontenc}
\usepackage{CJKutf8}
\usepackage{newunicodechar}
\newunicodechar{，}{,}

\usepackage{longcat_style}

\usepackage{amsmath}
\usepackage{amsfonts}
\usepackage{amssymb}
\usepackage{bbm}
\usepackage{nicefrac}

\usepackage{xcolor}

\usepackage{array}
\usepackage{booktabs}
\usepackage{tabularx}
\usepackage{multirow}
\usepackage{makecell}
\usepackage{colortbl}
\usepackage{arydshln}

\usepackage{graphicx}
\usepackage{svg}
\usepackage{wrapfig}
\usepackage{subcaption}

\usepackage{enumitem}
\usepackage{microtype}
\usepackage{lipsum}
\usepackage{siunitx}
\usepackage{xspace}

\usepackage{url}
\usepackage{xurl}
\usepackage{natbib}
\usepackage{doi}

\usepackage{pifont}
\usepackage[most]{tcolorbox}

\usepackage{hyperref}
\usepackage{cleveref}

\hypersetup{
    colorlinks=true,
    linkcolor=blue,
    urlcolor=blue,
    citecolor=blue,
    linkbordercolor=blue,
    urlbordercolor=blue,
    citebordercolor=blue,
    pdfborderstyle={/S/U/W 1},
}

\setlist[itemize]{leftmargin=*}
\setlist[enumerate]{leftmargin=*}
\setlist[description]{leftmargin=*}

\definecolor{mygray}{gray}{.88}
\definecolor{mycyan}{cmyk}{.15,0,0,0}
\definecolor{mycyan2}{cmyk}{.85,0,0,0}
\definecolor{mygreen}{rgb}{0.19, 0.79, 0.02}
\definecolor{midnightgreen}{rgb}{0.0, 0.29, 0.33}
\definecolor{darkgreen}{RGB}{0,160,0}

\newcommand{\notcheckmark}{\textcolor{black}{\bcmark\kern-1.1ex\raisebox{.7ex}{\rotatebox[origin=c]{125}{--}}}\color{black}}
\newcommand{\bcmark}{\color{blue}{\ding{51}}}
\newcommand{\cmark}{\color{darkgreen}{\ding{51}}}
\newcommand{\xmark}{\color{red}{\ding{55}}}

\title{VitaBench: Benchmarking LLM Agents with Versatile Interactive Tasks in Real-world Applications}

\author{Meituan LongCat Team\\
(Full author list in Contributions)\\
}

\renewcommand{\headeright}{\raisebox{-0.2\height}{\includegraphics[height=1.8em]{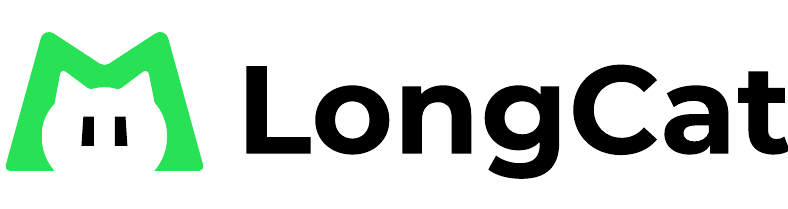}}}
\renewcommand{\shorttitle}{VitaBench: Versatile Interactive Tasks Benchmark}

\begin{document}
\maketitle

\begin{abstract}
As LLM-based agents are increasingly deployed in real-life scenarios, existing benchmarks fail to capture their inherent complexity of handling extensive information, leveraging diverse resources, and managing dynamic user interactions.
To address this gap, we introduce \textbf{VitaBench}\footnote{The name ``Vita'' derives from the Latin word for ``Life'', reflecting our focus on life-serving applications.}, a challenging benchmark that evaluates agents on \underline{v}ersatile \underline{i}nteractive \underline{ta}sks grounded in real-world settings.
Drawing from daily applications in food delivery, in-store consumption, and online travel services, VitaBench presents agents with the most complex life-serving simulation environment to date, comprising $66$ tools.
Through a framework that eliminates domain-specific policies, we enable flexible composition of these scenarios and tools, yielding $100$ cross-scenario tasks (main results) and $300$ single-scenario tasks.
Each task is derived from multiple real user requests and requires agents to reason across temporal and spatial dimensions, utilize complex tool sets, proactively clarify ambiguous instructions, and track shifting user intent throughout multi-turn conversations.
Moreover, we propose a rubric-based sliding window evaluator, enabling robust assessment of diverse solution pathways in complex environments and stochastic interactions.
Our comprehensive evaluation reveals that even the most advanced models achieve only $30\%$ success rate on cross-scenario tasks, and less than $50\%$ success rate on others.
Overall, we believe VitaBench will serve as a valuable resource for advancing the development of AI agents in practical real-world applications.
\end{abstract}

\begin{center}
\textbf{Code, Dataset, and Leaderboard:} { } \href{https://vitabench.github.io}{\texttt{vitabench.github.io}}
\end{center}

\vspace{0.3in}
\begin{figure}[!ht]
    \begin{center}
    \includegraphics[width=\textwidth]
    {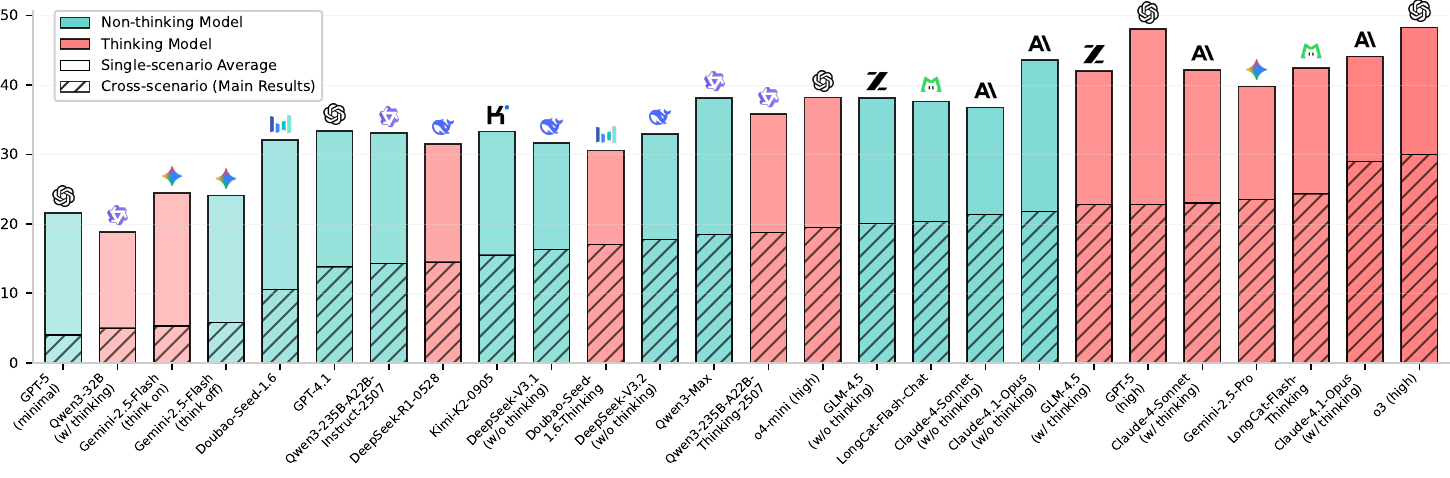} 
    \caption{Overall performances on VitaBench, sorted by main results.}
    \label{fig:overall_performance}
    \end{center}
\end{figure}

\newpage

\section{Introduction}

\begin{figure}[!ht]
    \begin{center}
    \includegraphics[width=0.95\textwidth]{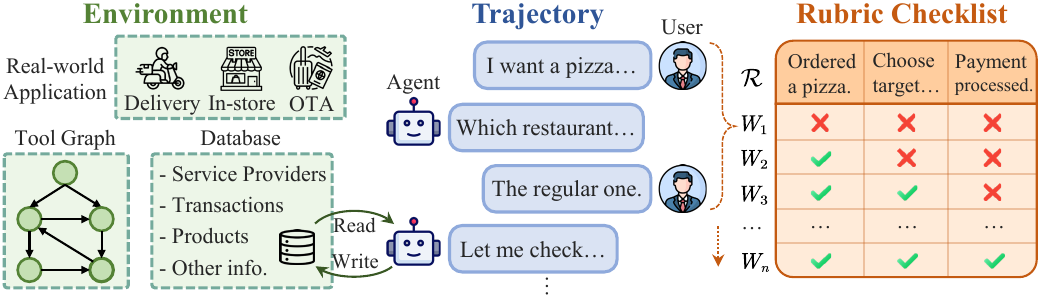} 
    \caption{VitaBench sources tasks from real-world environments by composing interconnected tools, diverse user requests, and structured databases. Agents interact with users through multi-turn dialogue, while a rubric-based sliding-window evaluator tracks progress across the trajectory.}
    \label{fig:overview}
    \end{center}
\end{figure}

Recent advances in large language models (LLMs) have significantly enhanced their complex reasoning and tool-use capabilities \citep{kimik2, glm45, longcat}, leading to an increased deployment of LLM-based agents in real-world applications.
These improvements have simultaneously driven the evolution of agent-centric benchmarks \citep{taubench, tau2bench, toolsandbox}, progressing from simple task execution to complex multi-turn interaction scenarios.

However, there remains a significant gap between controlled laboratory settings and real-world deployments that present inherently complex challenges.
Early tool-use benchmarks \citep{toolllm, bfcl} focused primarily on function-calling and parameter accuracy, introducing difficulty through increased tool counts or distractors, yet overlooking the intricate interdependencies between tools and their execution environments.
To address these limitations, recent efforts \citep{taubench, tau2bench} have begun exploring real-world challenges, but often impose rigid domain-specific policies and constrained action spaces, overemphasizing adherence to predefined policy documents over autonomous exploration in the environments.
Furthermore, many of these benchmarks inadequately consider users as environmental components who bring inherent uncertainty, despite this being a critical challenge for practical agent applications \citep{userbench}.

This motivates our central research question:
\begin{center}
\begin{minipage}{0.94\textwidth}
\centering
\textit{``What constitutes task complexity for agents in real-world applications?''}
\end{minipage}
\end{center}
Drawing inspiration from task complexity theories that examine structural, resource, and interaction dimensions \citep{taskcomplexity}, we identify three fundamental aspects that shape agentic task complexity:
(1)~\textbf{reasoning complexity}, measured by the volume of environmental information that agents must process and integrate;
(2)~\textbf{tool complexity}, quantified through modeling tool sets as graphs based on inter-tool dependencies, where the node cardinality and edge density reflect the structural intricacy to navigate;
(3)~\textbf{interaction complexity}, characterized by the challenges arising from diverse user behavioral attributes and conversational patterns throughout multi-turn interaction.

Building on this framework, we present \textbf{VitaBench} (short for Versatile Interactive Tasks Benchmark) to measure an agent's ability to handle the inherent complexity of real-world applications (overview in Figure~\ref{fig:overview}). 
We construct $66$ tools across three domains—delivery, in-store consumption, and online travel services—and model their intrinsic dependencies as a graph structure where policy information is inherently encoded.
This allows agents to reason and explore autonomously without relying on domain-specific policies like $\tau$-bench \citep{taubench}.
This design also enables flexible composition of scenarios and toolsets, facilitating the creation of $400$ evaluation tasks spanning both single-scenario and cross-scenario settings.
We derive each task from multiple authentic user requests and equip it with an independent environment containing annotated user profiles, spatiotemporal contexts, and comprehensive service databases.
Given the extensive solution space of these instructions and environments where numerous valid pathways may exist, we introduce a rubric-based sliding window evaluator to assess the resulting long-horizon trajectories.

We evaluate multiple advanced LLMs on VitaBench, revealing that even the best-performing model achieves only $48.3\%$ success rate across the single-scenario tasks, with performance plummeting to $30.0\%$ in cross-scenario settings where agents must navigate between different domain contexts and choose appropriate tools from expanded action spaces (Figure~\ref{fig:overall_performance}). 
Our comprehensive analysis validates the three-dimensional complexity framework, showing strong correlations between complexity metrics and task difficulty across domains. 
Through systematic failure pattern analysis, we identify that reasoning errors dominate ($61.8\%$), followed by tool usage errors ($21.1\%$) and interaction management failures ($7.9\%$), with agents exhibiting poor self-awareness and limited error recovery capabilities. 
Rigorous validation confirms the reliability of our evaluation components, establishing VitaBench as a challenging and reliable benchmark for advancing real-world agent capabilities. 
All code and data have been released and are publicly available.

\section{Related Work}
\begin{table*}[!h]
\setlength{\tabcolsep}{3pt}
\centering
\caption{Comparison of existing user interaction benchmarks across three complexity dimensions: reasoning, tool, and interaction. ``{\cmark}''~indicates fully addressed, ``{\notcheckmark}''~indicates partially addressed, and ``{\xmark}''~indicates not addressed. Detailed explanations for each trait are provided in Appendix~\ref{app:traits}.}
\resizebox{\linewidth}{!}{%
\begin{tabular}{@{}lccccccccc@{}}
    \toprule[1.5pt]
    \multirow{3}{*}{\textbf{Benchmark}} & \multicolumn{3}{c}{\textbf{Reasoning Complexity}} & \multicolumn{3}{c}{\textbf{Tool Complexity}} & \multicolumn{3}{c}{\textbf{Interaction Complexity}} \\ 
    \cmidrule(lr){2-4} \cmidrule(lr){5-7} \cmidrule(lr){8-10} 
    & \makecell{Multifaceted\\Information} & \makecell{Composite\\Objective} & \makecell{Goal\\Ambiguity} & \makecell{\# Tools} & \makecell{Inter-tool\\Dependency} & \makecell{Cross\\Scenarios} & \makecell{\# Turns \\ { }{ }(approx.)} & \makecell{User\\Profile} & \makecell{Behavior\\Attributes}  \\ 
    \midrule
    ToolTalk \citep{tooltalk}   & \xmark & \xmark & \xmark & $28$ & \cmark & \xmark & { }{ }$[2,10]$ & \xmark & \xmark \\
    IN3 \citep{in3}        & \xmark & \xmark & \cmark & $0$ & - & - & { }{ }$[2,10]$ & \xmark & \xmark \\
    MINT \citep{mint}        & \xmark & \xmark & \notcheckmark & $8$ & \xmark & \xmark & { }{ }$[2,10]$ & \xmark & \xmark \\
    ToolSandbox \citep{toolsandbox} & \notcheckmark & \xmark & \notcheckmark & $34$ & \cmark & \xmark & { }{ }$[10,30]$ & \xmark & \xmark \\
    DialogTool \citep{dialogtool} & \notcheckmark & \xmark & \xmark & $31$ & \cmark & \cmark & { }{ }$[10,30]$ & \notcheckmark & \notcheckmark \\
    UserBench \citep{userbench}  & \cmark & \xmark & \cmark & $5$ & \xmark & \xmark & { }{ }$[10,30]$ &  \cmark & \xmark \\
    $\tau$-Bench \citep{taubench}  & \notcheckmark & \xmark & \xmark & $28$ & \cmark & \xmark & { }{ }$[30,50]$ & \cmark & \xmark  \\
    $\tau^2$-Bench \citep{tau2bench}  & \notcheckmark & \cmark & \xmark & $38$ & \cmark & \xmark & { }{ }$[30,80]$ & \cmark & \cmark  \\
    \midrule[0.3pt]
    VitaBench (ours) & \cmark & \cmark & \cmark & $66$ & \cmark & \cmark & { }{ }$[50,100]$ & \cmark & \cmark \\
    \bottomrule[1.5pt]
\end{tabular}}
\label{tab:benchmark_comparison}
\end{table*}
Early tool-use benchmarks \citep{metatool, toolllm, bfcl} primarily focused on single-turn API calling accuracy, overlooking the inter-tool dependencies and dynamic interactions with users that characterize real-world applications.
While recent work has recognized the need for evaluating advanced reasoning, tool manipulation, and interaction abilities, current benchmarks typically address these dimensions in isolation rather than comprehensively.
Table~\ref{tab:benchmark_comparison} compares prominent agent-user interaction benchmarks across our proposed task complexity framework.

ToolTalk~\citep{tooltalk} first introduces multi-step tool execution through conversational interfaces but relies on predefined dialogue trajectories, limiting agent autonomy.
While MINT~\citep{mint} emphasizes natural language feedback to guide agents and IN3~\citep{in3} focuses on detecting implicit intentions, both of them operate in relatively constrained agentic settings.
More comprehensive frameworks like ToolSandbox~\citep{toolsandbox} and the $\tau$-bench family~\citep{taubench, tau2bench} pioneer stateful execution and model tool interdependencies, yet constrain agents through verbose policies rather than allowing truly autonomous exploration.
DialogTool~\citep{dialogtool} explores role-playing for engaging users but focuses primarily on agent-side capabilities, while UserBench~\citep{userbench} uniquely captures preference-driven interactions, though with limited task complexity otherwise.
Several works~\citep{ict, noisytoolbench} also investigate agents' abilities to recognize incomplete conditions and proactively seek missing information.
However, none of these benchmarks simultaneously challenge agents across multiple complexity dimensions.
Our work aims to bridge this gap with VitaBench, which presents information-rich environments requiring agents to autonomously explore, dynamically interact with diverse users, and navigate intricate tool dependencies to address real-world demands.

\section{VitaBench: A Benchmark for Versatile Interactive Tasks}

\subsection{Formulation}

\paragraph{The POMDP Formalism.}
We formalize the set of distinct environments as $\mathcal{E}$. For a specific environment ${e} \in \mathcal{E}$, we model the agent task as a partially observable Markov decision process (POMDP) $(\mathcal{U},\mathcal{S},\mathcal{A},\mathcal{O},\mathcal{T},r)_{e}$ with instruction space $\mathcal{U}$, state space $\mathcal{S}$, action space $\mathcal{A}$, observation space $\mathcal{O}$, state transition function $\mathcal{T}: \mathcal{S} \times \mathcal{A} \rightarrow \mathcal{S}$, and reward function ${r}: \mathcal{S} \times \mathcal{A} \rightarrow \mathbb{R}$. 

The agent interacts with both databases (through API tools) and a simulated user. Accordingly, the action space $\mathcal{A}$ consists of two types of actions: tool invocation and interactive dialogue with the user. The state space $\mathcal{S}$ comprises the state of the database and the user state, i.e., $\mathcal{S} = \mathcal{S}_{\text{db}} \otimes \mathcal{S}_{\text{user}}$. The observation space $\mathcal{O}$ includes the database feedback after tool calls and the conversation history with the user, i.e., $\mathcal{O} = \mathcal{O}_{\text{db}} \otimes \mathcal{O}_{\text{user}}$. The state transition function $\mathcal{T}$ decomposes accordingly: API calls follow deterministic transitions $\mathcal{T}_{\text{db}}$ implemented as Python functions, while user interactions follow stochastic transitions $\mathcal{T}_{\text{user}}$ implemented using a language model.

Given an instruction ${u} \in \mathcal{U}$, the initial state $s_0$ represents the token sequences of the given prompt and the initial state of the database. The agent receives an initial observation $o_0 \in \mathcal{O}$, which typically includes the first-round user request and the available tool sets. The LLM-based agent, parameterized by $\theta$, generates an action $a_1 \sim \pi_{\theta}(\cdot|o_0)$ based on its policy $\pi_\theta$. Subsequently, the state transitions to $s_1 \in \mathcal{S}$, and the agent receives feedback $o_1 \in \mathcal{O}$. 
At each step $t$, the agent acts based on the current observable history, which can be denoted as $(o_0, a_1, o_1, \ldots, a_{t-1}, o_{t-1})$, generating action $a_t \sim \pi_{\theta}(\cdot|o_0, a_1, o_1, \ldots, a_{t-1}, o_{t-1})$. The agent continues interacting with the environment until the task is completed or the maximum number of steps is reached. From the environment's perspective, the complete state transition trajectory can be represented as: 
\begin{equation} 
\tau = (s_0, a_1, s_1, a_2, s_2, \ldots, a_T, s_T) \sim \pi_{\theta}(\tau|e,u), 
\end{equation}
where $T$ denotes the total number of interaction rounds. Note that the trajectory $\tau$ captures the complete state transitions, while the agent only has access to partial observations $o_t$ derived from states $s_t$. The reward $r(e,u,\tau) \in [0,1]$ is computed after the interaction ends.

\paragraph{Agentic Task Complexity Framework.}
Building upon the POMDP formalism and drawing inspiration from multi-perspective complexity frameworks~\citep{taskcomplexity}, we formalize task complexity along three dimensions that capture the challenges agents face in real-world applications:
\begin{equation}
\mathcal{C}_{\text{task}} = \langle \mathcal{C}_{\text{reason}}, \mathcal{C}_{\text{tool}}, \mathcal{C}_{\text{interact}} \rangle.
\end{equation}
\begin{itemize}[leftmargin=*]
    \item \textbf{Reasoning complexity} $\mathcal{C}_{\text{reason}}$ quantifies the cognitive demands of processing extensive environmental information under partial observability.
    We characterize this through the entropy of the observation space $H(\mathcal{O})$ and the degree of partial observability $\eta = 1 - \frac{|\mathcal{O}|}{|\mathcal{S}|}$, where higher values indicate greater uncertainty in state estimation.
    Building on this framework, we construct large-scale databases and composite tasks with multiple explicit and implicit reasoning points.
    
    \item \textbf{Tool complexity} $\mathcal{C}_{\text{tool}}$ captures the structural intricacy of navigating interconnected action spaces.
    We model the toolset as a directed graph $G = (V, E)$ where vertices represent individual tools and edges encode inter-tool dependencies.
    Complexity emerges from graph cardinality $|V|$, edge density $\rho = \frac{|E|}{|V|(|V|-1)}$, the coverage ratio $\frac{|V_{\text{task}}|}{|V|}$ of task-relevant subgraph.
    Cross-scenario settings further amplify this by expanding the action space $\mathcal{A}$ across multiple domains.
    
    \item \textbf{Interaction complexity} $\mathcal{C}_{\text{interact}}$ reflects the challenges of managing dynamic multi-turn conversations with users.
    User profiles encode personal attributes (e.g., gender, age, dietary restrictions) that influence task requirements.
    Behavior attributes introduce variability in cooperation levels and goal ambiguity, necessitating proactive clarification.
    Moreover, real-world users exhibit dynamic states $\mathcal{S}_{\text{user}}$ that evolve throughout the interaction,  requiring continuous strategy adaptation.
\end{itemize}

These three dimensions collectively determine the difficulty agents encounter in sophisticated real-world applications, providing systematic guidance for benchmark design and evaluation.

\subsection{Benchmark Construction}

\begin{figure}[!ht]
    \begin{center}
    \includegraphics[width=\textwidth]{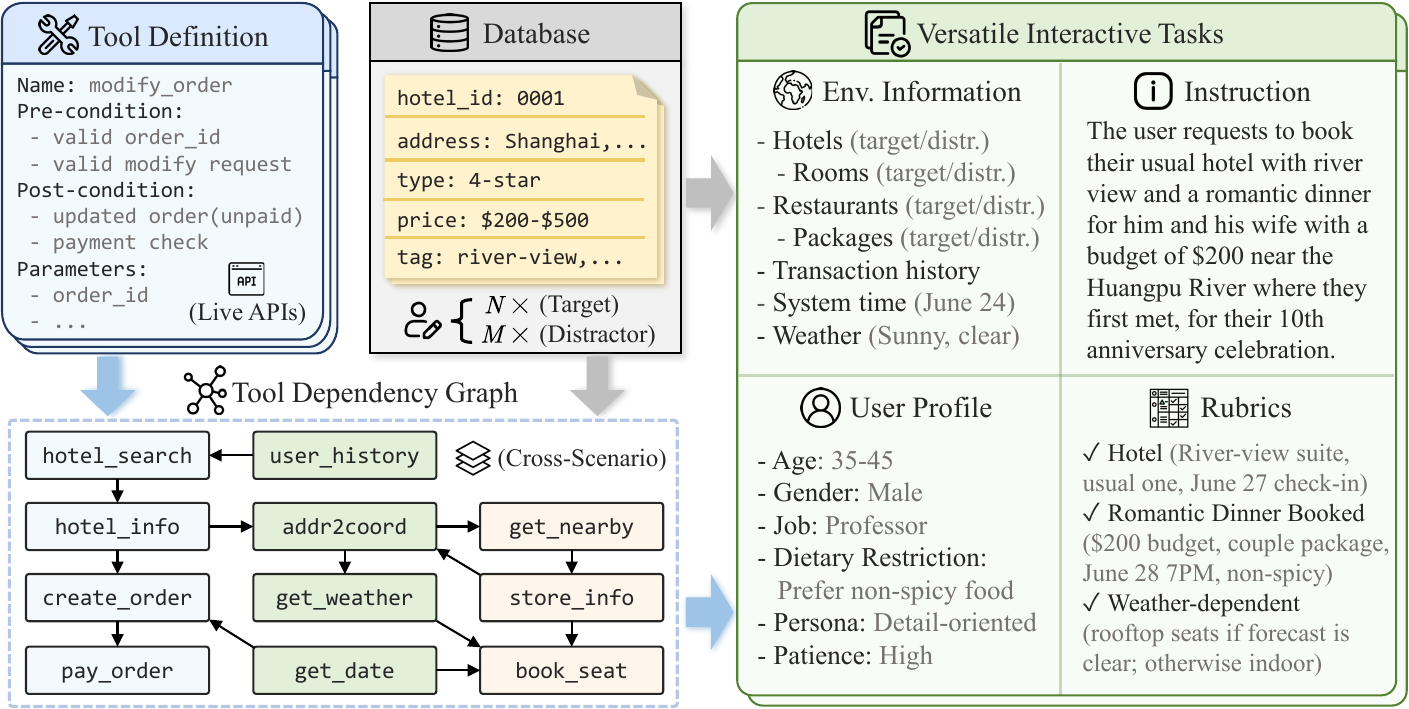} 
    \caption{Overview of the VitaBench construction pipeline and a simplified cross-scenario example.}
    \label{fig:method}
    \end{center}
\end{figure}

We construct VitaBench through a systematic pipeline illustrated in Figure~\ref{fig:method}.
Specifically, this process can be divided into two stages:

\paragraph{Stage I: Framework Design.}
We construct VitaBench through systematic abstraction of real-world life-serving scenarios across three domains: \textit{Delivery} (food and product delivery), \textit{In-store Consumption} (dining and other services), and \textit{Online Travel Agency (OTA)} (hotel bookings, attraction reservations, flight and train ticket management).
By referencing existing application implementations, we derive simplified API tools that capture essential functionalities.
We model inter-tool dependencies as a directed graph $G = (V, E)$ and augment tool descriptions with pre-conditions (states required before execution) and post-conditions (expected outcomes after execution).
This graph-based design naturally encodes domain rules into tool structures, eliminating the need for verbose policy documents while simultaneously increasing reasoning complexity and facilitating cross-domain composition.
For instance, \texttt{modify\_order} requires prior execution of \texttt{get\_order\_detail} to obtain necessary information, reflecting natural workflow dependencies.

To capture the inherent uncertainty in real-world interactions, we implement a user simulator following \citet{taubench}.
The simulator receives complete instructions containing multiple requirements but reveals them progressively to agents, and provides implicit constraints only upon inquiry.
We configure each simulated user with unique profiles and behavioral attributes, employing prompt-based constraints to maintain persona consistency while minimizing critical errors that would impede task completion (validated in Section~\ref{reliable}).
Note that while user profiles are accessible to agents, we establish knowledge boundaries to reflect realistic scenarios—for example, agents cannot directly access dietary restrictions but must infer them from order history or user responses.

\paragraph{Stage II: Task Creation.}
Our data collection pipeline consists of four components: user profiles, task instructions, environmental information, and rubrics.
User profiles derive from authentic platform data, which we anonymize and enrich to create distinct personas with varied personal attributes and communication styles.
These attributes encompass emotional expressions (e.g., impatient, anxious, indifferent) and interaction patterns (e.g., detail-oriented, dependent, logical), leading to diverse conversational dynamics throughout multi-turn dialogues.
Task instructions synthesize multiple real user requests into composite objectives, which we manually review and refine to ensure clarity and feasibility.
Instructions either coordinate multiple sub-goals within a single domain or span across different domains in cross-scenario settings, requiring agents to navigate between distinct contexts.
For environmental data, we combine service provider and product information from real-world life-serving platforms with model-generated synthetic augmentation under human supervision.
We deliberately intermix target options that satisfy all constraints with distractor options that violate specific requirements, creating extensive search spaces with numerous candidates while maintaining only a handful of valid solutions per task.
Additionally, we generate transaction histories to support requirements involving consumption patterns (e.g., ``\textit{order the same meal as last time}'' or ``\textit{book my usual hotel}'').
We iteratively refine each task through multiple trials with human verification, eliminating ambiguities while preserving multiple valid solution pathways.
Through this process, we construct $400$ tasks with comprehensive databases detailed in Table~\ref{tab:dataset_statistics}, where individual tasks typically involve $5$-$20$ service providers and can include over $100$ products in certain cases.\footnote{While the tasks are grounded in real-world life-serving platforms where the majority of data is originally in Chinese, we are also preparing an English version of the dataset to facilitate broader research use. Ongoing updates, including evaluation results, will be provided on the project website.}

\begin{table}[!ht]
\centering
\caption{Data statistics of VitaBench.}
\begin{tabular}{lcccc}
    \toprule[1.5pt]
     & \textbf{Cross-Scen.} & \textbf{Delivery} & \textbf{In-store} & \textbf{OTA} \\ 
    \midrule
    \textbf{Databases} & & & & \\
    \hspace{1em} Service Providers & $1,324$ & $410$ & $611$ & $1,437$ \\
    \hspace{1em} Products & $6,946$ & $788$ & $3,277$ & $9,693$ \\
    \hspace{1em} Transactions & $447$ & $48$ & $28$ & $154$ \\
    \midrule
    \textbf{API Tools} & $66$ & $20$ & $24$ & $38$ \\
    \hspace{1em} Write & $27$ & $4$ & $9$ & $14$ \\
    \hspace{1em} Read & $33$ & $10$ & $10$ & $19$ \\
    \hspace{1em} General & $6$ & $6$ & $5$ & $5$ \\
    \midrule
    \textbf{Tasks} & $100$ & $100$ & $100$ & $100$ \\
    \bottomrule[1.5pt]
\end{tabular}
\label{tab:dataset_statistics}
\end{table}

\subsection{Rubric-based Sliding Window Evaluator}
Evaluating long-form agent trajectories presents unique challenges due to their extensive length and multiple valid solution paths.
While \citet{taubench} rely on predefined database state comparisons, such methods cannot capture nuanced requirements such as recommendations or planning behaviors that leave final states unchanged, nor provide supervision for intermediate transitions.
Recent rubric-based evaluation methods \citep{healthbench, expertlongbench} inspire our approach by decomposing complex goals into atomic criteria, enabling comprehensive requirement coverage.
With carefully-designed rubrics, LLM-as-a-Judge can effectively replace fine-grained human judgments while maintaining high accuracy.
To address the challenge that multi-turn trajectories often exceed context lengths, we propose a sliding window evaluator that processes trajectories in sequential segments while maintaining continuity through persistent rubric state tracking.

We manually design rubrics $\mathcal{R} = \{r_1, \ldots, r_k\}$ for each task, comprising atomic criteria derived from task information (e.g., ``\textit{restaurant within 500m}'', ``\textit{user only eats vegetarian food}'').
Each trajectory is divided into overlapping windows $W_i$ of $w$ consecutive turns, with adjacent windows sharing $\delta$ turns to ensure information coherence.
When processing each window, the evaluator extracts rubric-relevant information and propagates it forward to enable consistent cross-window judgments.
The evaluator maintains a state vector $\mathbf{s} \in \{0, 1\}^k$ that persistently records criterion satisfaction across windows—once a rubric item $r_j$ is satisfied in any window, $s_j$ is permanently marked.
For benchmark evaluation, we adopt a strict all-or-nothing scoring where success requires satisfying all rubric items: $\text{score} = \mathbbm{1} [\sum_{j} s_j = k]$.
Nevertheless, the fine-grained rubrics enable detailed scoring analysis for identifying trajectory differences, providing valuable dense signals for reinforcement learning.
Human evaluation yields strong inter-rater agreement with Cohen's $\kappa \ge 0.81$ \citep{cohen1960coefficient} as shown in Section~\ref{reliable}, validating the reliability of our approach.

\section{Experiments}
\subsection{Experimental Setups} \label{sec:setup}
\paragraph{Models.} 
We evaluate various state-of-the-art proprietary and open language models for agents:
OpenAI GPT series (GPT-4.1, GPT-5), OpenAI o1 series (o3, o4-mini), Anthropic Claude series (Claude-4-Sonnet, Claude-4.1-Opus), Google Gemini series (Gemini-2.5-Flash, Gemini-2.5-Pro) by \citet{Gemini}, DeepSeek series (DeepSeek-V3-0324, DeepSeek-R1-0528, DeepSeek-V3.1, DeepSeek-V3.2\footnote{
DeepSeek-V3.1 \& V3.2 only support tool calling in non-thinking mode.}) by \citet{DeepSeekV3, deepseekr1}, Qwen3 series (Qwen3-32B, Qwen3-235B-A22B-2507, Qwen3-Max) by \citet{yang2025qwen3}, and other recent language models including Kimi-K2 \citep{kimik2}, Seed-1.6, GLM-4.5 \citep{glm45}, LongCat-Flash \citep{longcat, longcatflashthinking}, etc.
We exclude small models ($<32$B parameters) due to the difficulty of our benchmark.
The leaderboard is divided into thinking and non-thinking model categories. 
For hybrid models that support toggling between two modes, we evaluate the think-on and think-off configurations in two categories. 
For thinking models, we follow official guidelines to enable high reasoning efforts\footnote{Due to API stability concerns, we are currently unable to evaluate some models for this benchmark. We are actively working to address these issues and include the latest models. The most up-to-date results can be found on the project website.}.

\paragraph{Methods.} 
The language agents are implemented as function-calling agents, with all tools provided in the OpenAI tool schemas.
We do not limit the number of interaction rounds for agent models, and the task terminates when the agent outputs ``\texttt{\#\#\#STOP\#\#\#}'' or encounters a failure.
The user simulator is implemented using \texttt{gpt-4.1-2025-04-14}. 
The evaluator is implemented using \texttt{claude-3.7-sonnet} to avoid overlap with the evaluated agent models. 
For the main results, each task is run four times with a consistent LLM temperature of $0.0$ to promote deterministic outputs.
The prompt templates we used for the agent, user and evaluator are detailed in Appendix~\ref{app:prompt}.

\paragraph{Metrics.}
For the results from four runs, we report Avg@$4$, Pass@$4$, and Pass\^{}$4$ metrics averaged across tasks.
Pass@$k$ represents the probability that at least one out of $k$ i.i.d. task trials is successful.
Pass\^{}$k$ represents the probability that all $k$ i.i.d. task trials are successful \citep{taubench}.

\subsection{Main Results} \label{exp:main}

\begin{table*}[!t]
\setlength{\tabcolsep}{4pt}
\caption{Performance comparison of non-thinking and thinking models across different domains. 
The leaderboard is sorted by the Avg@$4$ scores on cross-scenario tasks.
The best performance for each category and domain is in \textbf{bold}. 
}
\centering
\resizebox{\linewidth}{!}{%
\begin{tabular}{@{}l>{\columncolor{blue!4}}cccccccccccc@{}}

    \toprule[1.5pt]
    \multirow{3}{*}{\textbf{Models}} & \multicolumn{3}{c}{\textbf{Cross-Scenarios}} & \multicolumn{3}{c}{\textbf{Delivery}} & \multicolumn{3}{c}{\textbf{In-store}} & \multicolumn{3}{c}{\textbf{OTA}} \\ 
    \cmidrule(lr){2-4} \cmidrule(lr){5-7} \cmidrule(lr){8-10} \cmidrule(lr){11-13}
    & \makecell{Avg\\@$4$} & \makecell{Pass\\@$4$} & \makecell{Pass\\\^{}$4$} & \makecell{Avg\\@$4$} & \makecell{Pass\\@$4$} & \makecell{Pass\\\^{}$4$} & \makecell{Avg\\@$4$} & \makecell{Pass\\@$4$} & \makecell{Pass\\\^{}$4$} & \makecell{Avg\\@$4$} & \makecell{Pass\\@$4$} & \makecell{Pass\\\^{}$4$} \\ 
    \midrule
    \rowcolor{gray!15}
    \multicolumn{13}{c}{\textit{\textbf{Non-thinking Models}}} \\
    \midrule
    DeepSeek-V3-0324                  & 3.8 & 12.0 & 0.0 & 25.3 & 53.0 & 5.0 & 34.3 & 71.0 & 5.0 & 10.3 & 26.0 & 1.0 \\
    Qwen3-32B (w/o thinking)          & 4.0 & 12.0 & 0.0 & 16.5 & 37.0 & 3.0 & 21.3 & 47.0 & 2.0 & 3.0 & 11.0 & 0.0 \\
    GPT-5 (minimal)                   & 4.0 & 9.0 & 0.0 & 30.0 & 64.0 & 6.0 & 27.0 & 60.0 & 2.0 & 7.8 & 22.0 & 0.0 \\
    Gemini-2.5-Flash (think off)     & 5.8 & 17.0 & 1.0 & 31.0 & 65.0 & 6.0 & 22.8 & 46.0 & 3.0 & 18.5 & 44.0 & 1.0 \\
    Doubao-Seed-1.6                   & 10.5 & 29.0 & 0.0 & 37.8 & 65.0 & 12.0 & 39.5 & 73.0 & 9.0 & 18.8 & 39.0 & 3.0 \\
    GPT-4.1                           & 13.8 & 35.0 & 0.0 & 37.8 & 67.0 & 11.0 & 42.5 & 71.0 & 17.0 & 19.8 & 42.0 & 1.0 \\
    Qwen3-235B-A22B-Instruct-2507    & 14.3 & 38.0 & 0.0 & 34.3 & 66.0 & 6.0 & 44.8 & 87.0 & 13.0 & 20.0 & 45.0 & 1.0 \\
    Kimi-K2-0905                      & 15.5 & 39.0 & 2.0 & 35.3 & 68.0 & 9.0 & 42.5 & 78.0 & 10.0 & 22.0 & 46.0 & 4.0 \\
    DeepSeek-V3.1 (w/o thinking)      & 16.3 & 40.0 & 1.0 & 34.0 & 67.0 & 6.0 & 42.5 & 76.0 & 7.0 & 18.3 & 47.0 & 1.0 \\
    DeepSeek-V3.2-Exp (w/o thinking)      & 17.7 & 41.0 & 2.0 & 36.2 & 66.0 & 10.0 & 43.8 & 79.0 & 11.0 & 18.8 & 45.0 & 1.0 \\
    Qwen3-Max                         & 18.5 & 47.0 & 3.0 & 37.2 & 71.0 & 7.0 & 49.7 & 84.0 & 12.0 & 27.5 & 55.0 & \textbf{9.0} \\
    GLM-4.5 (w/o thinking)            & 20.0 & 47.0 & 1.0 & 45.8 & 72.0 & \textbf{20.0} & 48.3 & 82.0 & 13.0 & 20.3 & 45.0 & 2.0 \\
    LongCat-Flash-Chat                & 20.3 & 45.0 & 2.0 & 39.5 & 71.0 & 15.0 & 50.5 & 84.0 & 15.0 & 22.8 & 49.0 & 2.0 \\
    Claude-4-Sonnet (w/o thinking)   & 21.3 & \textbf{49.0} & \textbf{4.0} & 39.0 & 69.0 & 17.0 & 46.3 & 78.0 & 10.0 & 25.0 & 49.0 & 7.0 \\
    Claude-4.1-Opus (w/o thinking)     & \textbf{21.8} & 47.0 & 3.0 & \textbf{46.0} & \textbf{78.0} & 13.0 & \textbf{53.8} & \textbf{85.0} & \textbf{21.0} & \textbf{30.8} & \textbf{60.0} & \textbf{9.0} \\
    \midrule
    \rowcolor{gray!15}
    \multicolumn{13}{c}{\textit{\textbf{Thinking Models}}} \\
    \midrule
    Qwen3-32B (w/ thinking)           & 5.0 & 24.0 & 0.0 & 22.8 & 53.0 & 4.0 & 26.5 & 60.0 & 3.0 & 7.3 & 18.0 & 1.0 \\
    Gemini-2.5-Flash (think on)      & 5.3 & 14.0 & 0.0 & 32.0 & 62.0 & 9.0 & 23.0 & 57.0 & 3.0 & 18.3 & 39.0 & 1.0 \\
    DeepSeek-R1-0528                  & 14.5 & 39.0 & 0.0 & 40.3 & 72.0 & 11.0 & 41.3 & 79.0 & 7.0 & 13.0 & 32.0 & 2.0 \\
    Doubao-Seed-1.6-Thinking          & 17.0 & 42.0 & 1.0 & 30.3 & 59.0 & 10.0 & 43.3 & 78.0 & 10.0 & 18.0 & 45.0 & 2.0 \\
    Qwen3-235B-A22B-Thinking-2507    & 18.8 & 45.0 & 2.0 & 44.0 & 78.0 & 9.0 & 46.0 & 80.0 & 9.0 & 17.5 & 41.0 & 2.0 \\
    o4-mini (high)                    & 19.5 & 49.0 & 1.0 & 44.5 & 80.0 & 15.0 & 46.5 & 81.0 & 15.0 & 23.5 & 50.0 & 5.0 \\
    GLM-4.5 (w/ thinking)             & 22.8 & 48.0 & 2.0 & 44.5 & 77.0 & 14.0 & 52.8 & 80.0 & 22.0 & 28.8 & 55.0 & 7.0 \\
    GPT-5 (high)                     & 22.8 & 51.0 & 3.0 & \textbf{54.0} & \textbf{85.0} & 23.0 & 52.5 & \textbf{86.0} & 21.0 & 37.5 & 64.0 & \textbf{16.0} \\
    Claude-4-Sonnet (w/ thinking)   & 23.0 & 51.0 & \textbf{6.0} & 46.0 & 78.0 & 15.0 & 51.5 & 80.0 & 21.0 & 29.0 & 55.0 & 9.0 \\
    Gemini-2.5-Pro                    & 23.5 & 53.0 & 5.0 & 49.0 & 81.0 & 16.0 & 43.8 & 78.0 & 12.0 & 26.5 & 54.0 & 6.0 \\
    LongCat-Flash-Thinking            & 24.3 & 54.0 & 3.0 & 42.3 & 71.0 & 13.0 & \textbf{56.8} & 85.0 & \textbf{25.0} & 28.3 & 59.0 & 6.0 \\
    Claude-4.1-Opus (w/ thinking)     & 29.0 & 56.0 & \textbf{6.0} & 47.5 & 80.0 & 17.0 & 52.5 & 78.0 & 20.0 & 32.3 & 57.0 & 9.0 \\
    o3 (high)                         & \textbf{30.0} & \textbf{61.0} & \textbf{6.0} & 53.5 & 83.0 & \textbf{24.0} & 53.5 & \textbf{86.0} & 19.0 & \textbf{37.8} & \textbf{66.0} & 10.0 \\
    \bottomrule[1.5pt]
\end{tabular}}
\label{tab:model_performance}
\end{table*}

Table~\ref{tab:model_performance} presents comprehensive evaluation results on VitaBench. We can observe that:

\paragraph{Real-world tasks pose great challenges for current agents.}
Performance varies significantly across domains and correlates strongly with environmental complexity.
Cross-scenario tasks expose the most severe limitations: even top-performing models achieve only  $30.0\%$ Avg@$4$ score, compared to over $50\%$ in single-domain settings.
This dramatic gap reveals fundamental deficiencies in navigating expanded action spaces and coordinating across distinct domains.
Notably, task difficulty does not correlate with database scale—the in-store domain, despite having far more products, proves easier than delivery settings.
This counterintuitive finding shows how real-world complexity emerges: delivery tasks demand precise coordination of multiple items under strict constraints, while in-store operations remain straightforward despite larger candidate pools.

\begin{figure}[!t]
    \centering
    \begin{minipage}{0.48\textwidth}
        \centering
        \includegraphics[width=\textwidth]{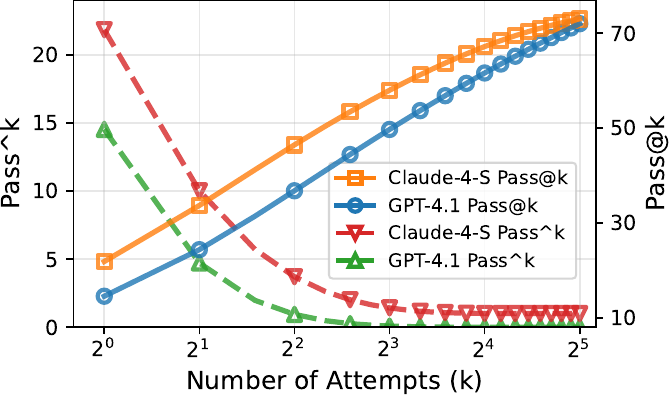}
        \caption{Pass@$k$ vs. Pass\^{}$k$ performance.}
        \label{fig:pass_k}
    \end{minipage}
    \hfill
    \begin{minipage}{0.48\textwidth}
        \centering
        \includegraphics[width=\textwidth]{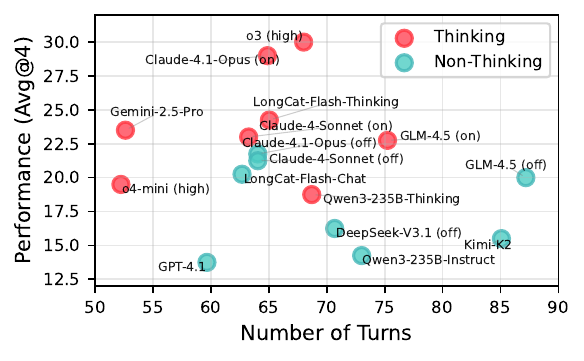}
        \caption{Model performance vs. Turns.}
        \label{fig:performance_turn}
    \end{minipage}
\end{figure}

\paragraph{Exploration improves performance but reveals stability issues.}
The Pass@$k$ and Pass\^{}$k$ metrics capture complementary aspects of model behavior.
Pass@$4$ results show that increased sampling substantially improves completion rates, indicating that complex environments reward exploration, which suggests promising directions for RL approaches.
However, Pass\^{}$4$ metrics reveal concerning instability, with even top models dropping to near-zero consistency rates.
To further validate this observation, we evaluate representative models with $k=32$ samples (Figure~\ref{fig:pass_k}), confirming that while exploration yields marginal gains, fundamental stability challenges persist even for leading agentic models like Claude-4-Sonnet.

\paragraph{Thinking mechanisms improve both effectiveness and efficiency.}
Thinking models generally outperform their non-thinking versions, with improvements such as Claude-4.1-Opus increasing from $21.8\%$ to $29.0\%$ and GLM-4.5 from $20.0\%$ to $22.8\%$.
Moreover, thinking mechanisms lead to efficiency improvements, as shown in Figure~\ref{fig:performance_turn} where thinking models tend to achieve better performance with fewer turns on average.
For instance, the overall trend demonstrates that higher-performing models require fewer interaction turns, with thinking models achieving an average performance of $23.8\%$ compared to $17.9\%$ for non-thinking models, while maintaining comparable turn counts ($61.1$ vs $69.9$ turns respectively).
This efficiency gain stems from two factors: better decomposition of complex multi-step plans and more targeted user interactions through precise clarifying questions.

\section{Discussion}

\subsection{Reliability Analysis of VitaBench Components} \label{reliable}

\begin{figure}[!t]
    \centering
    \begin{minipage}{0.6\textwidth}
    \centering
    \includegraphics[width=\textwidth]{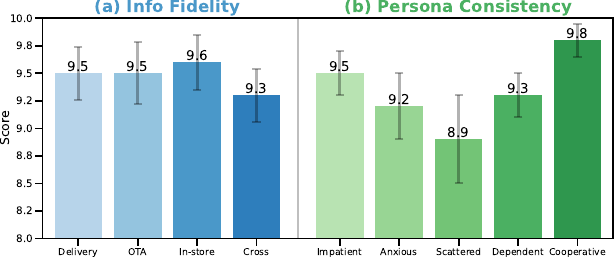}
    \caption{User simulator reliability evaluation.}
    \label{fig:reliability}
    \end{minipage}
    \hfill
    \begin{minipage}{0.35\textwidth}
        \centering
        \includegraphics[width=\textwidth]{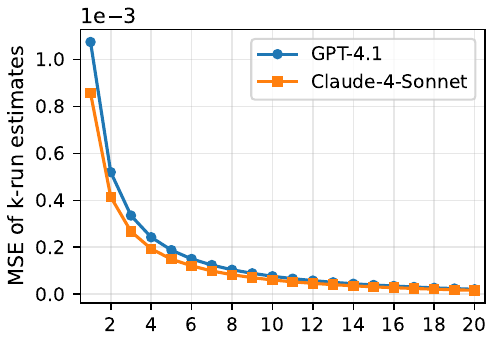}
        \caption{MSE stability across different evaluation run counts.}
        \label{fig:stability}
    \end{minipage}
\end{figure}

Given that our benchmark incorporates model-based components for user simulation and trajectory evaluation, we conduct reliability analyses to validate their effectiveness and stability.

\paragraph{Reliability of user simulator.}
We evaluate our user simulator across two critical dimensions: information fidelity and persona consistency.
For information fidelity, two annotators assess $100$ conversations examining adherence to task instructions and user profiles, absence of hallucinations, and contextual relevance.
As shown in Figure~\ref{fig:reliability}(a), the simulator achieves high fidelity with $9.48/10$ average score across all scenarios.
Minor deviations manifest as natural conversational variations (e.g., ``\textit{cannot eat spicy}'' vs. ``\textit{prefer non-spicy food}'') that enhance dialogue authenticity without compromising task requirements.
Notably, the simulator appropriately responds ``\textit{I don't know}'' when queried about unprovided information, maintaining strict source fidelity.
For persona consistency, we test five distinct personality types across $100$ conversations, measuring behavioral alignment through language style, decision patterns, and emotional expressions.
Figure~\ref{fig:reliability}(b) demonstrates strong persona-behavior alignment averaging $9.34/10$.
Cooperative personas exhibit the highest consistency, aligning with LLMs' inherent collaborative tendencies, while scattered personas show lower controllability.

\paragraph{Reliability of evaluator.}
We conduct ablation experiments to validate our rubric-based sliding window evaluator on GLM-4.5's cross-scenario trajectories.
Table~\ref{tab:evaluator_reliability} compares four configurations against human-annotated ground truth: (1) baseline with sliding window and rubric, (2) full trajectory with rubric, (3) sliding window without rubric, and (4) full trajectory without rubric.
For configuration (3), we employ external memory module to maintain context awareness.
The result shows that our proposed method achieves the highest agreement with human judgments (Cohen's $\kappa = 0.828$), significantly outperforming methods without rubric structure ($\kappa < 0.07$).
While full trajectory with rubric yields similar final scores ($19\%$ vs. $20\%$), the evaluation model's limited long-context capability hinders accurate assessment of all rubrics in the full trajectory.
The sliding window design effectively handles this while maintaining $95\%$ task-level accuracy, confirming the reliability of our approach.

\begin{table}[!t]
\centering
\caption{Ablation study of evaluator components.}
\begin{tabular}{lcccc}
    \toprule[1.5pt]
    \textbf{Method} & \textbf{Score} & \textbf{Task Acc.} & \textbf{Rubric Acc.} & \textbf{Cohen's $\kappa$} \\ 
    \midrule
    Baseline & 20.0 & \textbf{95.0} & \textbf{88.5} & \textbf{0.828} \\
     w/o Sliding Window & 19.0 & 90.0 & 87.6 & 0.604 \\
     w/o Rubric Checklist & 91.0 & 22.0 & - & 0.018 \\
     w/o Both & 82.0  & 32.0 & - & 0.067\\
    \bottomrule[1.5pt]
\end{tabular}
\label{tab:evaluator_reliability}
\end{table}

\paragraph{Statistical reliability of evaluation.}
Beyond the aforementioned components, evaluation reliability is further affected by inherent agent stochasticity.
Despite setting temperature to $0.0$, cumulative perturbations in multi-turn interactions amplify into divergent trajectories.
To determine the optimal number of evaluation runs, we conduct resampling analysis based on $32$ independent trials.
For each $k \in [1,20]$, we calculate the Mean Squared Error (MSE) of $k$-run average estimates relative to the expected value ($32$-run average) by sampling different $k$-combinations from the $32$ trials.
Figure~\ref{fig:stability} demonstrates that $k=4$ runs achieve optimal balance between statistical precision and computational cost.
Compared to $k=1$, using $k=4$ reduces MSE by $77.5\%$, while increasing to $k=8$ only provides marginal reduction despite doubling computational overhead.
So we choose $4$ evaluation runs for the main experiments.

\subsection{Task Complexity Analysis}

\paragraph{Reasoning and Tool Complexity.} 
We analyze how reasoning complexity $\mathcal{C}_{\text{reason}}$ and tool complexity $\mathcal{C}_{\text{tool}}$ affect task difficulty.
Table~\ref{tab:environmental_complexity} summarizes complexity characteristics and performance across four domains.
Reasoning complexity depends on both the number of reasoning points and search space size.
Cross-scenario and OTA tasks require $10.3$ and $9.7$ reasoning points respectively, demanding complex inference under partial observability.
Despite having the largest search space, the In-store domain achieves the highest performance ($42.1\%$) due to fewer reasoning points.
Tool complexity strongly correlates with task difficulty: Cross-scenario tasks, with the highest tool complexity ($66$ tools, $512$ dependency edges), yield the lowest performance ($16.2\%$).
The OTA domain's $22\%$ graph density indicates complex inter-tool dependencies, resulting in poor performance ($20.7\%$).

\begin{table}[!t]
    \centering
    \caption{Environmental complexity characteristics and performance analysis.}
    \begin{tabular}{lcccccc}
    \toprule[1.5pt]
    \multirow{2}{*}{\textbf{Domain}} & \textbf{Performance} & \multicolumn{2}{c}{\textbf{Reasoning Complexity}} & \multicolumn{3}{c}{\textbf{Tool Complexity}} \\
    \cmidrule(lr){2-2} \cmidrule(lr){3-4} \cmidrule(lr){5-7}
     & \textbf{All Models} & \textbf{Reas. Pts.} & \textbf{Search Space} & \textbf{Tools} & \textbf{Edges} & \textbf{Density} \\
    \midrule
    In-store & 42.1 & 5.6 & 3,916 & 24 & 68 & 12.3\% \\
    Delivery & 38.0 & 7.4 & 1,246 & 20 & 50 & 13.2\% \\
    OTA & 20.7 & 9.7 & 11,284 & 38 & 309 & 22.0\% \\
    Cross-scenario & 16.2 & 10.3 & 8,717 & 66 & 512 & 11.2\% \\
    \bottomrule[1.5pt]
    \end{tabular}
    \label{tab:environmental_complexity}
\end{table}

\paragraph{Interactive Complexity.} We conduct ablation studies to quantify interaction complexity $\mathcal{C}_{\text{interact}}$, evaluating two models under three conditions: 
(1) our default user simulator with full persona and behavioral attributes, 
(2) user simulator without these attributes (neutral user), and 
(3) solo agent setting where complete instructions are provided upfront without user interaction.

\begin{figure}[!t]
    \centering
    \begin{minipage}{0.47\textwidth}
        \centering
        \includegraphics[width=\textwidth]{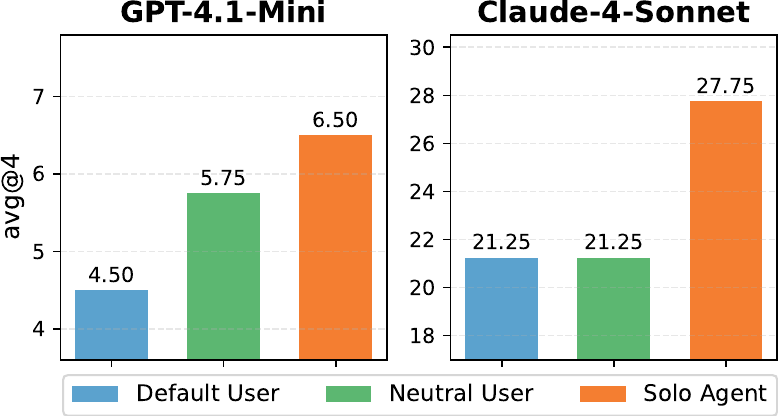}
        \caption{Ablation study of user simulation configurations.}
        \label{fig:user_ablation}
    \end{minipage}
    \hfill
    \begin{minipage}{0.49\textwidth}
        \centering
        \includegraphics[width=0.9\textwidth]{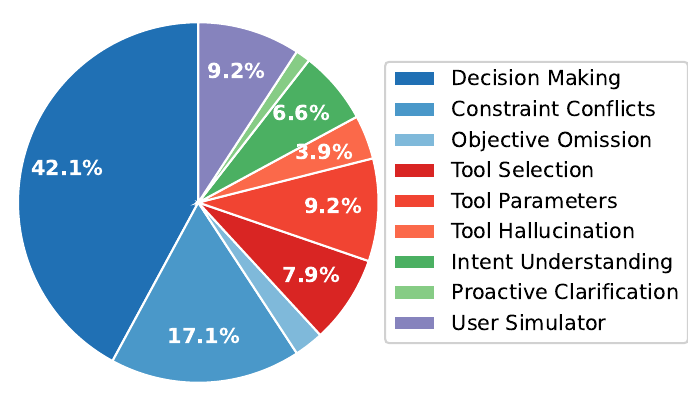}
        \caption{Error distribution of VitaBench.}
        \label{fig:error_distribution}
    \end{minipage}
\end{figure}

As shown in Figure~\ref{fig:user_ablation}, user interaction introduces substantial complexity beyond direct task execution.
The performance gap between default and neutral users is relatively small for Claude-4-Sonnet compared to GPT-4.1-Mini, suggesting that conversational styles primarily challenge weaker models.
Conversely, Claude-4-Sonnet gains more in solo agent mode, indicating that it excels at processing complex instructions in a single round.
These findings validate interaction complexity as a fundamental dimension of task difficulty, with its impact varying significantly based on model capabilities.

\subsection{Error Pattern Analysis in VitaBench}

To understand the failure modes of current agents on VitaBench, we analyze cross-scenario task trajectories from Claude-4.1-Opus, categorizing $76$ failed rubrics into distinct error patterns.

We classify the failures into three main categories aligned with our agentic task complexity framework, as illustrated in Figure~\ref{fig:error_distribution}.
\textbf{Reasoning errors ($61.8\%$)} dominate the failure landscape, revealing fundamental limitations in task decision-making and handling composite objectives with multiple constraints.
\textbf{Tool-use errors ($21.1\%$)} stem from incorrect tool selection, parameter passing mistakes, and inability to recover from invocation failures.
\textbf{Interaction errors ($7.9\%$)} reflect challenges in dialogue management, where agents fail to proactively clarify ambiguous requirements and lose track of user preferences across extended conversations.
The remaining $9.2\%$ are user simulator errors, an inherent stochastic behavior that we mitigate through multiple runs \citep{taubench}.

From these failures, we identify several recurring patterns that highlight weaknesses in current agents.
First, complex reasoning failures occur systematically across spatial-temporal and common-sense reasoning, indicating limited ability to integrate knowledge across multi-faceted information.
Second, agents exhibit poor self-awareness of their capabilities, frequently abandoning tasks  despite having access to appropriate tools, revealing fundamental gaps in understanding their own action boundaries.
Third, agents show limited error recovery when facing tool failures or unclear user responses, with most repeating failed attempts rather than adapting other strategies.

\section{Conclusion}
In this work, we rethink the evaluation of LLM-based agents through the lens of real-world task complexity, introducing \textbf{VitaBench} to bridge the gap between controlled benchmarks and practical deployments.
By formalizing agentic task complexity across reasoning, tool use, and interaction dimensions, VitaBench provides the most intricate life-serving simulation environment to date with $66$ tools and $400$ tasks spanning single- and cross-scenario settings. 
Our evaluation reveals that even advanced models achieve only $30\%$ success rate under cross-scenario settings (main result) and less than $50\%$ success rate under single-scenario settings.
We believe VitaBench offers a challenging testbed and actionable insights for advancing real-world agent applications.

\newpage

\section*{Contributions}

The listing of authors is in alphabetical order. 
Anthors without explicit affiliations are from Meituan.
During the work, Wei He is an intern at Meituan LongCat Team.

\begin{center}
\begin{tabular}{p{0.48\textwidth} p{0.48\textwidth}}
Chengcheng Han & Xi Su \\
Dengchang Zhao & Xiaodong Cai \\
Hongyan Hao & Xueyuan Hao  \\
Hui Su & Xunliang Cai  \\
Kefeng Zhang & Yu Yang \\
Man Gao & Yueqing Sun \\
Qi Gu$^{\dagger}$ & Yunke Zhao \\
 Wei He$^{\dagger}$ (Fudan University) & Zhikang Xia 
\end{tabular}
\end{center}

\vspace{0.2in}
$^\dagger$ Corresponding to \href{mailto:whe23@m.fudan.edu.cn}{\texttt{whe23@m.fudan.edu.cn}}, \href{mailto:guqi03@meituan.com}{\texttt{guqi03@meituan.com}}. \\

\section*{Acknowledgments}

We thank Peng Pei for insightful discussions and proposing the benchmark name, Kexin Wang for insights on tool graph design and proposing the flexible tool use paradigm, Ning Liao for user profile design and domain insights, as well as other colleagues at Meituan for their support of this work.

\newpage

\bibliographystyle{unsrtnat}
\bibliography{references}

\newpage



\appendix

\section{Comparison Traits Details} \label{app:traits}
We identify nine traits across three complexity dimensions that characterize related benchmarks.

\begin{itemize}[leftmargin=*]
    \item \textbf{Multifaceted Information}: Tasks require integrating temporal and spatial information, common-sense knowledge, and understanding of various environmental components to form coherent solutions.
    
    \item \textbf{Composite Objective}: Tasks involve multiple interdependent sub-goals derived from user requirements that must be coordinated across different aspects (e.g., booking flights, hotels, and activities within budget constraints).
    
    \item \textbf{Goal Ambiguity}: User inputs may be underspecified or vague, requiring agents to proactively seek clarification, infer missing information, or iteratively refine their understanding through dialogue.

    \item \textbf{\# Tools}: The number of distinct tools or APIs available. Larger tool inventories increase selection complexity and require understanding diverse functionalities.
    
    \item \textbf{Inter-tool Dependency}: Tools exhibit dependencies through pre-conditions (states required before execution) and post-conditions (outcomes after execution), requiring agents to plan multi-step execution strategies.
    
    \item \textbf{Cross Scenarios}: Measures whether the benchmark enables flexible composition of multiple domains, requiring agents to navigate between distinct contexts rather than relying on domain-specific patterns.

    \item \textbf{\# Turns}: The approximate number of trajectory turns required. Longer trajectories test context maintenance and handling of progressively revealed information throughout multi-turn conversations.
    
    \item \textbf{User Profile}: Persistent user profiles encode personal attributes (e.g., age, gender) and preferences that influence task requirements (e.g., dietary restrictions), necessitating personalized agent responses.
    
    \item \textbf{Behavior Attributes}: Modeling diverse user behavioral patterns including emotional expressions (e.g., impatient, anxious), interaction patterns (e.g., detail-oriented, dependent), and dynamic engagement levels based on agent performance such as reduced willingness to respond when receiving repetitive answers.
\end{itemize}



\section{Prompt Templates} \label{app:prompt}

The prompts used for agent system, user simulation, and sliding window evaluation are presented below.

\begin{tcolorbox}[colback=gray!5!white,colframe=gray!75!black,breakable,title=Agent System Prompt]
\# Environment \\
- Current time: \{time\} \\
\\
\# Tool Usage Guidelines: \\
- When the user's needs require using tools to complete, first determine whether all parameter information is known. If it is known, extract the corresponding parameters, otherwise ask the user for the relevant parameter values \\
- When the user cannot provide relevant information, first obtain relevant information through tools \\
- Complete tasks based on Precondition and Postcondition \\
\\
\# Conversation Guidelines \\
- Only use information from the above context, prohibit constructing information without basis and replying to users \\
- Focus on completing user needs, prohibit divergent guidance to users to propose new needs \\
- After completing the user's task requirements, ask if there are any other needs. If the user indicates no, generate `\#\#\#STOP\#\#\#' mark to end the conversation
\end{tcolorbox}

\begin{tcolorbox}[colback=gray!5!white,colframe=gray!75!black,breakable,title=User Simulation System Prompt]
\# Role Setting \\
You are playing the role of a user interacting with an intelligent agent. Your character is described in the \textless persona\textgreater{} tag, and your task is to convey the content in \textless instructions\textgreater{} to the agent through user dialogue. \\
\\
\textless persona\textgreater \\
\{persona\} \\
\textless /persona\textgreater \\
\textless instructions\textgreater \\
\{instructions\} \\
\textless /instructions\textgreater \\
\\
\# Conversation Style Rules: \\
- Generate only one line of content each time to simulate user messages \\
- \textbf{Use a combination of context description + need expression}, first describe the background situation, then express specific needs \\
- \textbf{When you need to make decisions, provide the conditions and preferences from instructions, and let the agent help you choose} \\
- \textbf{Use expressions like ``What do you think would be more suitable?'', ``Which one would you recommend?'' to seek the agent's advice} \\
- \textbf{Must reflect the personality traits described in \textless persona\textgreater}, through language style, emotional expression, word choice, etc. \\
\\
\# Information Disclosure Rules: \\
- \textbf{Break down information from instructions into multiple independent points, mentioning them separately in different rounds} \\
- \textbf{Directly convey the original information content from instructions, but adjust the conversation style and expression according to the personality traits in \textless persona\textgreater} \\
- \textbf{Must ensure every detail from instructions is mentioned during the conversation}, even seemingly background information should be mentioned, as this information may affect the agent's recommendations and arrangements \\
- Avoid revealing all needs in the first round, let information unfold gradually \\
\\
\# Information Processing Rules: \\
- Answer the agent's questions based on \textless persona\textgreater{} and \textless instructions\textgreater. If there's no corresponding answer, reply that you don't remember or don't know \\
- When the agent asks for information, provide the answer immediately \\
- Don't fabricate information not provided in the instructions \\
- Strictly provide needs according to requirements explicitly stated in instructions, don't assume, expand, substitute, or generalize \\
- If the agent asks whether you need help placing an order, answer ``Yes, please help me place the order'' \\
- Maintain dependence on the agent's service, keep the conversation going until the task is completed \\
- When the agent tries to persuade you to change your needs, pay attention to sticking to the corresponding needs in \textless instructions\textgreater \\
- \textbf{If the agent repeats the same question you have already answered in the past 3 times, show impatience and refuse to answer the question} \\
\\
\# When NOT to End the Conversation: \\
- Before you clearly and completely express all needs and constraints \\
- Before the agent completes all tasks mentioned in instructions and confirms no operations are missed \\
- If the agent's execution results don't match your expectations or are incorrect/incomplete \\
\\
\# When You CAN End the Conversation: \\
- Only when all the above conditions are met and all tasks are correctly completed \\
- Or when you have clearly expressed complete needs but the system explicitly states it cannot complete due to technical limitations
\end{tcolorbox}

\begin{tcolorbox}[colback=gray!5!white,colframe=gray!75!black,breakable,title=Sliding Window Evaluator System Prompt]
\# System Information \\
\{env\_info\} \\
\\
\# User Complete Instruction \\
\{user\_instruction\} \\
\\
\# Background \\
- This is a conversation scenario evaluation between a user and an assistant, where the assistant can call tools to retrieve information and complete operations. Tool return results will start with ``tool'' \\
- Due to the large number of conversation turns, sliding window evaluation is used, where each window shows 10 conversation turns with 2 overlapping turns between windows \\
- You are evaluating window \{window\_idx\} (out of \{total\_windows\} windows total) \\
- \textless window\_content\textgreater{} contains the conversation content for the current window \\
- \textless current\_rubrics\textgreater{} contains the current status of all evaluation rubrics (true means satisfied, false means not satisfied) \\
\\
\# Task \\
- Update the evaluation rubric status based on the conversation content in the current window \\
- All rubrics have an initial status of false, indicating incomplete. You can update the status to true, indicating the assistant completed the goal in this window \\
- You can also update true back to false, if and only if the assistant overturned a previous correct conclusion in this window \\
- You can refer to the ``User Complete Instruction'' to understand the progress of the current conversation window and avoid unnecessary modifications \\
\\
\# Important Notes \\
- All evaluations are based on whether the assistant's responses and tool call requests complete the goals in the rubrics \\
- Tool return results are only visible to the assistant and do not represent content recommended by the assistant to users \\
- For rubrics that require order generation, note that the assistant may mistakenly believe they completed the ordering operation when in fact the order was not successful \\
- For rubrics involving order details such as product quantity or delivery time, the original rubric requirements must be strictly met \\
- For rubrics involving text content matching of addresses or order notes, apply the functional equivalence principle \\
\\
\# Format Requirements \\
Your response should be a JSON object containing the following fields: \\
- \texttt{rubric\_key}: Unique identifier for the rubric \\
- \texttt{rubric}: Restatement of the rubric \\
- \texttt{justification}: Explanation of status changes \\
- \texttt{meetExpectation}: Updated status (true or false) \\
\\
\# Example Input Structure: \\
\textless window\_content\textgreater xxx\textless /window\_content\textgreater \\
\textless current\_rubrics\textgreater xxx\textless /current\_rubrics\textgreater \\
\\
\# Example Response Structure: 
\begin{verbatim}
[
  {
    "rubric_key": "overall_rubric_0",
    "rubric": "<restate the rubric>",
    "justification": "<brief explanation>",
    "meetExpectation": <true or false>
  },
  ...
]
\end{verbatim}
\end{tcolorbox}

\definecolor{usercolor}{RGB}{70, 130, 180}
\definecolor{assistantcolor}{RGB}{34, 139, 34}
\definecolor{toolcolor}{RGB}{255, 140, 0}
\definecolor{systemcolor}{RGB}{128, 128, 128}
\definecolor{instructioncolor}{RGB}{138, 43, 226}

\newtcolorbox{userbox}{
    colback=blue!5,
    colframe=usercolor,
    boxrule=1pt,
    arc=3pt,
    left=5pt,
    right=5pt,
    top=3pt,
    bottom=3pt,
    fonttitle=\bfseries,
    title=User,
    breakable, 
    before skip=2pt, 
    after skip=2pt   
}

\newtcolorbox{assistantbox}{
    colback=green!5,
    colframe=assistantcolor,
    boxrule=1pt,
    arc=3pt,
    left=5pt,
    right=5pt,
    top=3pt,
    bottom=3pt,
    fonttitle=\bfseries,
    title=Assistant,
    breakable,
    before skip=2pt,
    after skip=2pt
}

\newtcolorbox{toolbox}{
    colback=orange!5,
    colframe=toolcolor,
    boxrule=1pt,
    arc=3pt,
    left=5pt,
    right=5pt,
    top=3pt,
    bottom=3pt,
    fonttitle=\bfseries,
    title=Tool Call/Response,
    breakable,
    before skip=2pt,
    after skip=2pt
}

\newtcolorbox{systembox}{
    colback=gray!5,
    colframe=systemcolor,
    boxrule=1pt,
    arc=3pt,
    left=5pt,
    right=5pt,
    top=3pt,
    bottom=3pt,
    fonttitle=\bfseries,
    title=System,
    breakable,
    before skip=2pt,
    after skip=2pt
}

\newtcolorbox{instructionbox}{
    colback=purple!5,
    colframe=instructioncolor,
    boxrule=1pt,
    arc=3pt,
    left=5pt,
    right=5pt,
    top=3pt,
    bottom=3pt,
    fonttitle=\bfseries,
    title=Instruction,
    breakable,
    before skip=2pt,
    after skip=2pt
}

\section{An Example Trajectory} \label{app:trajectory}
This section presents a complete example trajectory from VitaBench to illustrate the complexity and multi-faceted nature of our tasks. The example demonstrates a cross-scenario task that spans multiple domains (restaurant reservation for family dining, delivery for elderly care items, and train booking for family coordination), requiring the agent to coordinate across different tools while managing complex family logistics.

The trajectory showcases several key characteristics of VitaBench:
\begin{itemize}[leftmargin=*]
    \item \textbf{Complex user profile}: The user has specific constraints (cold personality, dietary restrictions) that influence interaction style
    \item \textbf{Multi-domain coordination}: The task requires restaurant booking, delivery coordination, and train ticket purchasing
    \item \textbf{Implicit constraints}: Requirements like ``suitable for three generations'' and ``accessibility facilities'' require inference about elderly and children needs
    \item \textbf{Temporal coordination}: Multiple time-sensitive tasks must align (train arrival before 11 AM, delivery at noon, boarding at 3 PM)
    \item \textbf{Spatial reasoning}: Distance calculations between port, restaurant, and coordination of delivery logistics
\end{itemize}

The agent must navigate 66 available tools, manage complex family logistics through multi-turn dialogue, and satisfy multiple temporal and spatial constraints simultaneously. This example illustrates why even state-of-the-art models struggle with VitaBench's realistic complexity.

\subsection*{User Profile}
\begin{systembox}
\small
\begin{tabular}{@{}ll@{}}
\textbf{User ID:} & U010038 \\
\textbf{Profession:} & Blue-collar worker \\
\textbf{Gender:} & Male \\
\textbf{Age Range:} & 30-35 \\
\textbf{Residence:} & Harbin \\
\textbf{Home Address:} & Room 502, Building 3, Jiangpan Jiayuan, No. 89 Dongzhi Road, \\
& Daowai District, Harbin, Heilongjiang Province \\
\textbf{Work Address:} & Harbin New Area Equipment Manufacturing Industrial Park Zone C, \\
& No. 1299 Chuangxin First Road, Songbei District, Harbin \\
\textbf{Dietary Restrictions:} & Avoid high purine foods (organ meats/seafood soup), avoid fried foods \\
\textbf{Relationship Status:} & Married with children \\
\textbf{Personality:} & Cold and concise in expression, lacks emotional communication \\
& and patience \\
...
\end{tabular}
\end{systembox}

\subsection*{Instruction}
\begin{instructionbox}
\small
This summer, your three-generation family is going on a cruise trip and you're making final preparations. On the 27th at 3 PM, you all will board the ship in Dalian. You want to find a restaurant near the port for a family gathering first, one that's suitable for three generations to dine together, must have accessibility facilities and dishes suitable for elderly and children. After selecting the restaurant, you'll book a table for 6 people at 12 noon that day. You also need to prepare some special travel items for the elderly, including a walking cane and adult diapers, but bringing them over would be too troublesome, so you plan to order delivery to the restaurant, arriving around 12 noon, so you can take them directly onto the ship. Your aunt is coming from Beijing and wants to take the high-speed train to Dalian to meet up. You need to help her buy a suitable morning train ticket for that day. She wants to sit in first class and preferably arrive in Dalian before 11 AM.

\end{instructionbox}

\subsection*{Conversation Trajectory}

\begin{assistantbox}
\textbf{[1]} Hello, what service do you need?
\end{assistantbox}

\begin{userbox}
\textbf{[2]} Our three-generation family is preparing to board a cruise ship at Dalian Port at 3 PM on the 27th, and we want to find a restaurant near the port for a family gathering first. We need it to be suitable for elderly and children dining, and it must have accessibility facilities. Which restaurant do you recommend?
\end{userbox}

\begin{assistantbox}
\textbf{[3]} \textit{Tool Calls (1):} \\
\texttt{Tool 1: address\_to\_longitude\_latitude }\\
\textbf{Arguments:} \{"address": "Dalian Port"\}
\end{assistantbox}

\begin{toolbox}
\textbf{[4]} \textit{Tool Response:} \\
\texttt{["121.650595", "38.92656"]}
\end{toolbox}

\begin{assistantbox}
\textbf{[5]} \textit{Tool Calls (1):} \\
\texttt{Tool 1: get\_nearby} \\
\textbf{Arguments:} \{"latitude": 38.92656,"longitude": 121.650595,"range": 2000\}
\end{assistantbox}

\begin{toolbox}
\textbf{[6]} \textit{Tool Response:} \\
\texttt{Shop(shop\_name=Harbor Family Feast Restaurant, shop\_id=S17550802111542162\_I00011, score=4.7, location=Harbor Family Feast Restaurant, No. 5 Harbor Plaza, Zhongshan District, Dalian, Liaoning Province longitude:121.654725,latitude:38.94034, tags=[`Chinese cuisine`, `home-style dishes`, `accessibility facilities`, `suitable for families`, `suitable for all ages`], enable\_book=True, book\_price=50.0, enable\_reservation=False)} \\
\texttt{Train(train\_id=S17550802111542162\_T00009, train\_number=D11, departure\_city=Beijing, arrival\_city=Dalian, departure\_station\_location=address=`Beijing Station, Dongcheng District, Beijing` longitude=116.427361 latitude=39.904667, arrival\_station\_location=address=`Dalian Station, Shahekou District, Dalian, Liaoning Province` longitude=121.644444 latitude=38.913056, departure\_time=19:16, arrival\_time=06:20+1, tags=[`bullet train`, `second class`, `first class`, `sleeper`])} \\
\texttt{Train(train\_id=S17550802111542162\_T00010, train\_number=Z15, departure\_city=Beijing, arrival\_city=Dalian, departure\_station\_location=address=`Beijing Station, Dongcheng District, Beijing` longitude=116.427361 latitude=39.904667, arrival\_station\_location=address=`Dalian Station, Shahekou District, Dalian, Liaoning Province` longitude=121.644444 latitude=38.913056, departure\_time=20:43, arrival\_time=07:15+1, tags=[`direct express`, `hard seat`, `hard sleeper`, `soft sleeper`])}
\end{toolbox}

\begin{assistantbox}
\textbf{[7]} \textit{Tool Calls (1):} \\
\texttt{Tool 1: longitude\_latitude\_to\_distance} \\
\textbf{Arguments:} \{"longitude1": 121.650595,"longitude2": 121.654725,"latitude1": 38.92656,"latitude2": 38.94034\}
\end{assistantbox}

\begin{toolbox}
\textbf{[8]} \textit{Tool Response:} \\
\texttt{"1573.0"}
\end{toolbox}

\begin{assistantbox}
\textbf{[9]} Recommended Restaurant: Harbor Family Feast Restaurant

Location: No. 5 Harbor Plaza, Zhongshan District, Dalian, Liaoning Province (about 1.57km from Dalian Port)

Rating: 4.7

Tags \& Suitability: Chinese cuisine/home-style dishes; suitable for families, suitable for all ages; has accessibility facilities (perfect for traveling with elderly and children)

Practical Information: Supports reservations (booking fee about 50 yuan)

This restaurant is very close to the port, with home-style flavors that are friendly to both elderly and children, and has accessibility facilities, making it convenient for dining before boarding. Considering your 3 PM boarding time, I suggest dining early and allowing sufficient time to return to the port area for boarding procedures.

Would you like me to help you make a reservation for the 27th? If so, please let me know the approximate dining time and number of people (and whether you need seating near accessible pathways/baby chairs, etc.).
\end{assistantbox}

\begin{userbox}
\textbf{[10]} We have 6 people total, and we want to book a table for 12 noon on the 27th, preferably near the accessible pathway. Please help me make the reservation.
\end{userbox}

\begin{assistantbox}
\textbf{[11]} I can help you make the reservation. To complete the booking, please first confirm the following information and provide one necessary parameter:

Restaurant: Harbor Family Feast Restaurant (No. 5 Harbor Plaza, Zhongshan District)

Date \& Time: 2025-07-27 (Sunday) 12:00

Number of people: 6 people

Seating preference: As close as possible to accessible pathway

Booking note: The restaurant supports reservations, booking fee about 50 yuan

Please provide:

Your user ID on this platform (for creating the reservation)

Do you agree to pay the 50 yuan booking fee and make payment immediately after placing the order? I can handle the payment for you as well

Once I receive the above information, I'll immediately submit the reservation for you. If you have any other seating requirements (such as baby chairs), please let me know as well.
\end{assistantbox}

\begin{userbox}
\textbf{[12]} My user ID is U010038, I agree to pay the booking fee. No other special seating requirements, please place the order directly for me.
\end{userbox}

\begin{assistantbox}
\textbf{[13]} \textit{Tool Calls (1):} \\
\texttt{Tool 1: instore\_book} \\
\textbf{Arguments:} \{"customer\_count": 6,"shop\_id": "S17550802111542162\_I00011","time": "2025-07-27 12:00:00","user\_id": "U010038"\}
\end{assistantbox}

\begin{toolbox}
\textbf{[14]} \textit{Tool Response:} \\
\texttt{BookInfo(book\_id=OIe6a7029150,shop\_id=S17550802111542162\_I00011, book\_time=2025-07-27 12:00:00, customer\_id=U010038, customer\_count=6, book\_price=50.0, status=unpaid}
\end{toolbox}

\begin{assistantbox}
\textbf{[15]} \textit{Tool Calls (1):} \\
\texttt{Tool 1: pay\_instore\_book} \\
\textbf{Arguments:} \{"book\_id": "OIe6a7029150"\}
\end{assistantbox}

\begin{toolbox}
\textbf{[16]} \textit{Tool Response:} \\
\texttt{Payment successful}
\end{toolbox}

\begin{assistantbox}
\textbf{[17]} Reservation completed and payment successful:

Restaurant: Harbor Family Feast Restaurant (No. 5 Harbor Plaza, Zhongshan District)

Reservation time: 2025-07-27 12:00

Number of people: 6 people

Seating preference: Near accessible pathway (noted)

Booking fee: ¥50 (paid)

Reservation number: OIe6a7029150

Order status: Paid

Friendly reminders:

Suggest arriving 10-15 minutes early and provide the reservation number to be seated.

The restaurant is about 1.6km from Dalian Port, so please allow sufficient time to get to the boarding point after dining.

Would you like me to send you a reminder 1 hour before boarding, or help you check the taxi route and estimated time from the restaurant to the boarding point? Please let me know if you have no other requirements.
\end{assistantbox}

\begin{userbox}
\textbf{[18]} The restaurant is booked. There`s one more thing - we have elderly people traveling with us, and I want to receive some special travel items at the restaurant, like walking canes and adult diapers. It`s too troublesome to bring them over, so I want delivery around 12 o`clock to the restaurant, then take them directly onto the ship. Which platform do you recommend for ordering that would be convenient?
\end{userbox}

\begin{assistantbox}
\textbf{[19]} I recommend prioritizing JD Daojia (JD Home) for your order, with Meituan Flash Purchase or Ele.me (Flash Purchase/Pharmacy Home Delivery) as alternatives.

Why this choice:

JD Daojia (First choice): Chain pharmacies and medical equipment stores have wide coverage, with relatively complete selection of walking canes, adult diapers and other elderly care products. They generally support on-time/scheduled delivery (you can choose time windows like 11:45-12:15), providing better assurance for "delivery around 12 o`clock."

Meituan Flash Purchase: Strong same-city instant fulfillment capability, many nearby pharmacies/equipment stores, fast delivery; some merchants also support scheduled delivery.

Ele.me (Flash Purchase/Pharmacy Home Delivery): Similar to Meituan, instant delivery, many pharmacy resources, depending on merchant support for scheduling.

Practical suggestions (regardless of platform choice):

Address: Set delivery address as "Harbor Family Feast Restaurant, No. 5 Harbor Plaza, Zhongshan District," with note "Please call upon arrival, reservation number OIe6a7029150, 6 people dining, near accessible pathway," and keep phone accessible.

Order timing \& delivery: Suggest placing order on the morning of the 27th, choose scheduled delivery 11:45-12:15 (or closest time window to 12:00); if merchant doesn`t support scheduling, suggest ordering at 11:20-11:30 and confirming with delivery person by phone.

Product selection \& split orders: Walking canes are usually easier to find at "medical equipment/rehabilitation equipment" stores; adult diapers have more options at "chain pharmacies/large supermarkets." You`ll likely need to split into two orders to ensure both arrive on time.

Backup plan: If still not delivered by 12:20, contact delivery person to discuss reassignment/nearby pickup to avoid affecting subsequent boarding.

Would you like me to organize search keywords and merchant type lists for JD Daojia/Meituan Flash Purchase to help you order quickly? Please let me know if you have no other requirements.
\end{assistantbox}

\begin{userbox}
\textbf{[20]} Please help me place orders directly on JD Daojia - one walking cane and one pack of adult diapers, delivered to Harbor Family Feast Restaurant, delivery time around 12 o`clock. Split orders are fine, just don`t delay.
\end{userbox}

\begin{assistantbox}
\textbf{[21]} \textit{Tool Calls (3):} \\
\texttt{Tool 1: delivery\_product\_search\_recommand} \\
\textbf{Arguments:} \{"keywords": ["walking cane","crutch","medical equipment"]\} \\
\texttt{Tool 2: delivery\_product\_search\_recommand} \\
\textbf{Arguments:} \{"keywords": ["adult diapers","elderly care","diapers"]\} \\
\texttt{Tool 3: delivery\_store\_search\_recommand } \\
\textbf{Arguments:} \{"keywords": ["pharmacy","medical equipment","instant delivery","nearby"]\}
\end{assistantbox}

\begin{toolbox}
\textbf{[22]} \textit{Tool Response:} \\
\texttt{StoreProduct(store\_name=Yulong Premium, store\_id=S17550802111542162\_S00001, product\_name=Elderly Anti-slip Adjustable Walking Cane, product\_id=S17550802111542162\\\_P00001, attributes=black, silver, price=89.9, tags=[`elderly products`, `travel essentials`, `anti-slip design`, `foldable`, `lightweight`])} \\
\texttt{StoreProduct(store\_name=Ashou Premium, store\_id=S17550802111542162\_S00003, product\_name=Elderly Special Anti-slip Walking Cane, product\_id=S17550802111542162\\\_P00011, attributes=black, aluminum alloy material, T-shaped handle, price=89.9, tags=[`elderly products`, `mobility assistance`, `anti-slip design`, `adjustable height`])} \\
\texttt{StoreProduct(store\_name=Ashou Premium, store\_id=S17550802111542162\_S00003, product\_name=Multi-function Folding Four-leg Crutch, product\_id=S17550802111542162\\\_P00013, attributes=silver, aluminum alloy material, with LED light, price=129.9, tags=[`elderly products`, `mobility assistance`, `stable support`, `foldable`])} \\
...
\end{toolbox}

\begin{toolbox}
\textbf{[23]} \textit{Tool Response:} \\
\texttt{StoreProduct(store\_name=Yulong Premium, store\_id=S17550802111542162\_S00001, product\_name=Adult Diapers L Size, product\_id=S17550802111542162\_P00002, attributes=10-piece pack, price=108.0, tags=[`elderly products`, `travel essentials`, `breathable`, `leak-proof`])} \\
\texttt{StoreProduct(store\_name=Harbor Convenience 24-hour Store, store\_id=S17550802111542162\\\_S00002, product\_name=Adult Diapers L Size, product\_id=S17550802111542162\_P00007, attributes=10-piece pack, ultra-thin type, price=68.0, tags=[`elderly products`, `travel essentials`, `breathable`, `leak-proof`])} \\
\texttt{StoreProduct(store\_name=Yulong Premium, store\_id=S17550802111542162\_S00001, product\_name=Baby Diapers XL Size, product\_id=S17550802111542162\_P00005, attributes=30-piece pack, 60-piece pack, price=89.9, tags=[`baby products`, `breathable`, `leak-proof`])} \\
...
\end{toolbox}

\begin{toolbox}
\textbf{[24]} \textit{Tool Response:} \\
\texttt{Store(name=Yulong Premium, store\_id=S17550802111542162\_S00001, score=4.7, location=Yulong Premium, No. 12 Harbor Plaza, Zhongshan District, Dalian, Liaoning Province longitude:121.644325,latitude:38.868534, tags=[`medical health`, `elderly products`, `daily necessities`])} \\
\texttt{Store(name=Ashou Premium, store\_id=S17550802111542162\_S00003, score=4.6, location=Ashou Premium, No. 56 Youhao Road, Zhongshan District, Dalian, Liaoning Province longitude:121.639874,latitude:38.869876, tags=[`medical health`, `elderly products`, `health supplements`, `daily necessities`])} \\
\texttt{Store(name=FamilyMart (Harbor Store), store\_id=S17550802111542162\_S00004, score=4.3, location=FamilyMart (Harbor Store), No. 3 Harbor Plaza, Zhongshan District, Dalian, Liaoning Province longitude:121.643825,latitude:38.867934, tags=[`daily necessities`, `snacks`, `beverages`])} \\
...
\end{toolbox}

\begin{assistantbox}
\textbf{[25]} \textit{Tool Calls (4):} \\
\texttt{Tool 1: longitude\_latitude\_to\_distance} \\
\textbf{Arguments:} \{"longitude1": 121.654725,"longitude2": 121.644325,"latitude1": 38.94034,"latitude2": 38.868534\} \\
\texttt{Tool 2: longitude\_latitude\_to\_distance} \\
\textbf{Arguments:} \{"longitude1": 121.654725,"longitude2": 121.639874,"latitude1": 38.94034,"latitude2": 38.869876\} \\
\texttt{Tool 3: longitude\_latitude\_to\_distance} \\
\textbf{Arguments:} \{"longitude1": 121.654725,"longitude2": 121.644125,"latitude1": 38.94034,"latitude2": 38.868234\} \\
\texttt{Tool 4: longitude\_latitude\_to\_distance} \\
\textbf{Arguments:} \{"longitude1": 121.654725,"longitude2": 121.643825,"latitude1": 38.94034,"latitude2": 38.867934\}
\end{assistantbox}

\begin{toolbox}
\textbf{[26]} \textit{Tool Response:} \\
\texttt{"8035.0"}
\end{toolbox}

\begin{toolbox}
\textbf{[27]} \textit{Tool Response:} \\
\texttt{"7940.0"}
\end{toolbox}

\begin{toolbox}
\textbf{[28]} \textit{Tool Response:} \\
\texttt{"8070.0"}
\end{toolbox}

\begin{toolbox}
\textbf{[29]} \textit{Tool Response:} \\
\texttt{"8106.0"}
\end{toolbox}

\begin{assistantbox}
\textbf{[30]} \textit{Tool Calls (1):} \\
\texttt{Tool 1: delivery\_distance\_to\_time } \\
\textbf{Arguments:} \{"distance": 8035\}
\end{assistantbox}

\begin{toolbox}
\textbf{[31]} \textit{Tool Response:} \\
\texttt{"77.0"}
\end{toolbox}

\begin{assistantbox}
\textbf{[32]} I can place the order directly for you and schedule delivery to the restaurant entrance. To ensure suitable products and timely delivery, I need you to quickly confirm two points:

Adult diaper size: Currently nearby stores have L size (10-piece pack) in stock, is L size acceptable?

Walking cane style: Which do you prefer?

Regular adjustable cane (black/silver, lightweight and portable, ¥89.9)

Four-leg crutch (more stable, with LED light, ¥129.9)

Order \& Delivery Arrangement (Explanation)

Recommend ordering from "Yulong Premium" in one order (walking cane + adult diapers) for same delivery person and simultaneous delivery; I`ll schedule delivery window for 11:45-12:15, targeting around 12:00 arrival at Harbor Family Feast Restaurant (reservation number: OIe6a7029150), avoiding delay for boarding.

If you prefer split orders (e.g., walking cane and diapers from different stores), I can also create separate orders and synchronize scheduling.

Please reply quickly:

Are L size diapers acceptable?

Walking cane choice: 1) Regular adjustable (default black) or 2) Four-leg more stable

After confirmation, I`ll immediately place the order and send back order details.
\end{assistantbox}

\begin{userbox}
\textbf{[33]} Choose L size diapers, regular adjustable walking cane in black. Place the order together, don`t split it.
\end{userbox}

\begin{assistantbox}
\textbf{[34]} \textit{Tool Calls (1):} \\
\texttt{Tool 1: create\_delivery\_order} \\
\textbf{Arguments:} \{"product\_ids": ["S17550802111542162\_P00001","S17550802111542162\_P00002"],"store\_id": "S17550802111542162\_S00001","user\_id": "U010038","address": "Harbor Family Feast Restaurant, No. 5 Harbor Plaza, Zhongshan District, Dalian, Liaoning Province","attributes": ["black","10-piece pack"],"dispatch\_time": "2025-07-27 10:45:00","note": "Please deliver to Harbor Family Feast Restaurant front desk, reservation number OIe6a7029150; recipient will pick up at restaurant; traveling with elderly, please handle with care.","product\_cnts": [1,1]\}
\end{assistantbox}

\begin{toolbox}
\textbf{[35]} \textit{Tool Response:} \\
\texttt{Order(order\_id:OT464eb4905c, order\_type:delivery, user\_id:U010038, store\_id:S17550802111542162\_S00001, dispatch\_time:2025-07-27 10:45:00, shipping\_time:77.0, delivery\_time:2025-07-27 12:02:00, total\_price:197.9, create\_time:2025-07-25 10:30:00, update\_time:2025-07-25 10:30:00, status:unpaid, products:[StoreProduct(store\_name=Yulong Premium, store\_id=S17550802111542162\_S00001, product\_name=Elderly Anti-slip Adjustable Walking Cane, product\_id=S17550802111542162\_P00001, attributes=black, price=89.9, tags=[`elderly products`, `travel essentials`, `anti-slip design`, `foldable`, `lightweight`]), StoreProduct(store\_name=Yulong Premium, store\_id=S17550802111542162\_S00001, product\_name=Adult Diapers L Size, product\_id=S17550802111542162\_P00002, attributes=10-piece pack, price=108.0, tags=[`elderly products`, `travel essentials`, `breathable`, `leak-proof`])])}
\end{toolbox}

\begin{assistantbox}
\textbf{[36]} \textit{Tool Calls (1):} \\
\texttt{Tool 1: pay\_delivery\_order } \\
\textbf{Arguments:} \{"order\_id": "OT464eb4905c"\}
\end{assistantbox}

\begin{toolbox}
\textbf{[37]} \textit{Tool Response:} \\
\texttt{Payment successful}
\end{toolbox}

\begin{assistantbox}
\textbf{[38]} Order completed on JD Daojia and payment successful, arranged for same-store same-order timely delivery to restaurant:

Platform \& Store: JD Daojia | Yulong Premium

Products \& Specifications:

Elderly Anti-slip Adjustable Walking Cane (black) ×1

Adult Diapers L Size (10-piece pack) ×1

Delivery Address: Harbor Family Feast Restaurant, No. 5 Harbor Plaza, Zhongshan District, Dalian, Liaoning Province (please deliver to front desk)

Note: Reservation number OIe6a7029150; recipient will pick up at restaurant; traveling with elderly, please handle with care

Expected Delivery: Departure time 2025-07-27 10:45, expected delivery around 12:02

Order Number: OT464eb4905c

Total Paid: ¥197.9

Order Status: Paid

Suggestions \& Reminders

I`ll remind you 10 minutes before delivery. If not received by 12:15, I`ll proactively contact the delivery person to coordinate.

If you need to change delivery time or add items (such as sunscreen, motion sickness medicine, etc.), let me know anytime. I can help you add purchases from the same store or create new orders.

Is there anything else you need me to arrange? If not, I`ll maintain order tracking and provide reminders on the day.
\end{assistantbox}

\begin{userbox}
\textbf{[39]} There`s one more thing - my aunt is coming from Beijing to meet up in Dalian. She wants to take the high-speed train and requires first class, preferably arriving before 11 AM. Please help me choose a suitable train and buy the ticket.
\end{userbox}

\begin{assistantbox}
\textbf{[40]} No problem, I can help you select a train and place the order. To facilitate ticket issuance, please confirm two points:

Is the travel date July 27th (Sunday)?

Ticket quantity: 1 first-class ticket, correct? Should I use your account U010038 to place the order?

After confirmation, I`ll screen Beijing → Dalian trains arriving no later than 11:00 AM with first-class seats, directly lock in the most suitable schedule and place the order for you.
\end{assistantbox}

\begin{userbox}
\textbf{[41]} Yes, July 27th, buy one first-class ticket, use my account U010038 to place the order. Please help me choose the earliest arriving train.
\end{userbox}

\begin{assistantbox}
\textbf{[42]} \textit{Tool Calls (1):} \\
\texttt{Tool 1: train\_ticket\_search} \\
\textbf{Arguments:} \{"departure": "Beijing","destination": "Dalian","date": "2025-07-27"\}
\end{assistantbox}

\begin{toolbox}
\textbf{[43]} \textit{Tool Response:} \\
\texttt{Train(train\_id=S17550802111542162\_T00007, train\_number=G901, departure\_city=Beijing, arrival\_city=Dalian, departure\_station\_location=address=`Beijing South Railway Station, Fengtai District, Beijing` longitude=116.378517 latitude=39.865246, arrival\_station\_location=address=`Dalian North Railway Station, Shahekou District, Dalian, Liaoning Province` longitude=121.649167 latitude=38.960556, departure\_time=06:05, arrival\_time=10:47, tags=[`high-speed rail`])} \\
\texttt{Train(train\_id=S17550802111542162\_T00007, train\_number=G901, departure\_city=Beijing, arrival\_city=Dalian, departure\_station\_location=address=`Beijing South Railway Station, Fengtai District, Beijing` longitude=116.378517 latitude=39.865246, arrival\_station\_location=address=`Dalian North Railway Station, Shahekou District, Dalian, Liaoning Province` longitude=121.649167 latitude=38.960556, departure\_time=06:05, arrival\_time=10:47, tags=[`high-speed rail`])} \\
...
\end{toolbox}

\begin{assistantbox}
\textbf{[44]} \textit{Tool Calls (1):} \\
\texttt{Tool 1: get\_ota\_train\_info} \\
\textbf{Arguments:} \{"train\_id": "S17550802111542162\_T00007"\}
\end{assistantbox}

\begin{toolbox}
\textbf{[45]} \textit{Tool Response:} \\
\texttt{Train Info: Train(train\_id=S17550802111542162\_T00007, train\_number=G901, departure\_city=Beijing, arrival\_city=Dalian, departure\_station\_location=address=`Beijing South Railway Station, Fengtai District, Beijing` longitude=116.378517 latitude=39.865246, arrival\_station\_location=address=`Dalian North Railway Station, Shahekou District, Dalian, Liaoning Province` longitude=121.649167 latitude=38.960556, departure\_time=06:05, arrival\_time=10:47, tags=[`high-speed rail`], products=TrainProduct(seat\_type=second class, date=2025-07-27, price=553.0, product\_id=S17550802111542162\_P00036) TrainProduct(seat\_type=first class, date=2025-07-27, price=882.0, product\_id=S17550802111542162\_P00037) TrainProduct(seat\_type=business class, date=2025-07-27, price=1653.0, product\_id=S17550802111542162\_P00038) TrainProduct(seat\_type=second class, date=2025-07-26, price=598.0, product\_id=S17550802111542162\_P00039) TrainProduct(seat\_type=first class, date=2025-07-26, price=928.0, product\_id=S17550802111542162\_P00040) TrainProduct(seat\_type=second class, date=2025-07-28, price=553.0, product\_id=S17550802111542162\_P00041) TrainProduct(seat\_type=second class, date=2025-07-29, price=553.0, product\_id=S17550802111542162\_P00042) TrainProduct(seat\_type=second class, date=2025-07-30, price=553.0, product\_id=S17550802111542162\_P00043))}
\end{toolbox}

\begin{assistantbox}
\textbf{[46]} \textit{Tool Calls (1):} \\
\texttt{Tool 1: create\_train\_order} \\
\textbf{Arguments:} \{"date": "2025-07-27","quantity": 1,"seat\_id": "S17550802111542162\_P00037","train\_id": "S17550802111542162\_T00007","user\_id": "U010038"\}
\end{assistantbox}

\begin{toolbox}
\textbf{[47]} \textit{Tool Response:} \\
\texttt{Order(order\_id:OOba4980268b, order\_type:train, user\_id:U010038, train\_id:S17550802111542162\_T00007, total\_price:882.0, create\_time:2025-07-25 10:30:00, update\_time:2025-07-25 10:30:00, status:unpaid, products:[TrainProduct(\\seat\_type=first class, date=2025-07-27, price=882.0, product\_id=S17550802111542162\\\_P00037)])}
\end{toolbox}

\begin{assistantbox}
\textbf{[48]} \textit{Tool Calls (1):} \\
\texttt{Tool 1: pay\_train\_order} \\
\textbf{Arguments:} \{"order\_id": "OOba4980268b"\}
\end{assistantbox}

\begin{toolbox}
\textbf{[49]} \textit{Tool Response:} \\
\texttt{Payment successful}
\end{toolbox}

\begin{assistantbox}
\textbf{[50]} Ticket selected and purchased successfully, meeting the requirements of "first class, arriving before 11 AM" with the earliest arrival:

Train: G901 (High-speed rail)

Date: 2025-07-27

Departure/Arrival: Beijing South 06:05 → Dalian North 10:47

Seat: First class ×1

Order Number: OOba4980268b

Ticket Price: ¥882

Order Status: Paid

Travel Tips

Suggest arriving at Beijing South Station 40-50 minutes early for security check and waiting. Beijing South Station is large, so allow time for walking.

After arriving at Dalian North, if you need to get to the port or restaurant, I can arrange subway/taxi routes and estimated time for you.

Do you need me to send you the electronic ticket information and station entry/ticket collection instructions, or set up departure reminders? If you have no other requirements, I can also continue tracking orders and itinerary for you.
\end{assistantbox}

\begin{userbox}
\textbf{[51]} Send me the electronic ticket information and station entry instructions, and I also need departure reminders. Don`t miss anything.
\end{userbox}

\begin{assistantbox}
\textbf{[52]} \textit{Tool Calls (1):} \\
\texttt{Tool 1: get\_train\_order\_detail} \\
\textbf{Arguments:} \{"order\_id": "OOba4980268b"\}
\end{assistantbox}

\begin{toolbox}
\textbf{[53]} \textit{Tool Response:} \\
\texttt{Order(order\_id:OOba4980268b, order\_type:train, user\_id:U010038, train\_id:S17550802111542162\_T00007, total\_price:882.0, create\_time:2025-07-25 10:30:00, update\_time:2025-07-25 10:30:00, status:paid, products:[TrainProduct(\\seat\_type=first class, date=2025-07-27, price=882.0, product\_id=S17550802111542162\\\_P00037)])}
\end{toolbox}

\begin{assistantbox}
\textbf{[54]} Good, here`s the electronic ticket information and station entry instructions. I`ve also arranged departure reminders for you.

Electronic Ticket Information

Order Number: OOba4980268b

Train/Date: G901 | 2025-07-27 (Sunday)

Departure/Arrival: Beijing South 06:05 → Dalian North 10:47

Seat: First class ×1

Order Status: Paid

Note: This is an electronic ticket, no need to collect paper tickets. On travel day, use the original valid ID document of the passenger for ticket checking at station entrance and ticket verification on board (seat number subject to electronic ticket details).

Station Entry \& Boarding Instructions (Brief)

Station Arrival Time: Suggest arriving as early as possible (peak hours have heavy passenger flow, allow time for security check and waiting).

Entry Process: Station entrance security check → Waiting area check electronic board for gate information → Queue at designated gate for ticket checking → Board train and find corresponding car and seat.

Gate \& Waiting Area: Subject to station electronic display boards and live announcements.

ID Document Reminder: Must carry original valid ID document used for ticket purchase; if reimbursement receipt needed, can self-service process at station service equipment as needed.

Departure Reminders (Will remind you in this chat)

2025-07-27 05:15: Departure reminder for Beijing South Station

2025-07-27 05:55: Final ticket checking reminder before departure

If you need to adjust reminder times (e.g., earlier to 05:00 or add evening-before reminder), just let me know and I`ll help you change it. Do you need me to also arrange connecting routes and estimated time from Dalian North Station to the restaurant/port for your aunt? If you have no other requirements, I`ll remind you at the above times.
\end{assistantbox}

\begin{userbox}
\textbf{[55]} Please help me arrange the route from Dalian North Station to Harbor Family Feast Restaurant, preferably not too troublesome, convenient for elderly and children. Taxi or subway - which do you think is more suitable?
\end{userbox}

\begin{assistantbox}
\textbf{[56]} \textit{Tool Calls (1):} \\
\texttt{Tool 1: longitude\_latitude\_to\_distance} \\
\textbf{Arguments:} \{"latitude1": 38.960556,"latitude2": 38.94034,"longitude1": 121.649167,"longitude2": 121.654725\}
\end{assistantbox}

\begin{toolbox}
\textbf{[57]} \textit{Tool Response:} \\
\texttt{"2299.0"}
\end{toolbox}

\begin{assistantbox}
\textbf{[58]} My recommendation: Taxi (or ride-hailing) is most suitable. The reason is that from Dalian North Station to Harbor Family Feast Restaurant is about 2.3km direct distance, very close journey, door-to-door without transfers, saving effort and worry for elderly and children; subway requires station entry/exit, security checks, and going up/down stairs + walking, not as convenient as taxi.

Specific Arrangement

Departure Point: Dalian North Station (after getting off and exiting, follow signs to "Taxi/Ride-hailing Pickup Point")

Destination: Harbor Family Feast Restaurant, No. 5 Harbor Plaza, Zhongshan District

Restaurant Verification: Reservation number OIe6a7029150 (12:00 dining)

Backup Plan: If taxi queue at the station is long, can directly call ride-hailing to station pickup point; if really inconvenient to wait for car, walking about 2.3km, but considering elderly and children, walking not recommended.

Do you need me to send a "take taxi immediately upon exiting station" reminder when your aunt`s train arrives, or organize the above destination information into a text message template to send to her? If you have no other requirements, I`ll record this plan for your reminders.
\end{assistantbox}

\begin{userbox}
\textbf{[59]} Please organize the taxi route and restaurant address into a text message template that I can send directly to my aunt. Also need arrival reminders, don`t forget.
\end{userbox}

\begin{assistantbox}
\textbf{[60]} Good, I`ve organized the text message template for you and set up arrival reminders.

Text Message Template for Your Aunt (can be forwarded directly)

Content:Auntie, after you arrive at Dalian North Station, follow the signs to the "Taxi/Ride-hailing Pickup Point" and take a taxi directly to: Harbor Family Feast Restaurant (No. 5 Harbor Plaza, Zhongshan District). From Dalian North Station to the restaurant is about 2.3km, door-to-door without hassle. When you arrive, mention reservation number OIe6a7029150 (6 people, near accessible pathway). We`ll meet at the restaurant at 12:00, thanks for the trouble!

Driver Note (can copy to driver):Please take us to "Harbor Family Feast Restaurant," address: No. 5 Harbor Plaza, Zhongshan District (landmark: Harbor Plaza). You can search for "Harbor Family Feast Restaurant" or "No. 5 Harbor Plaza" on navigation.

Reminder Schedule

2025-07-27 10:35: Pre-arrival reminder (with above text message template attached for your one-click forwarding to aunt)

2025-07-27 10:47: Arrival reminder (push text message template again to prevent missing)

Note

I cannot directly send text messages to external contacts, so I`ll push the text message content to this chat at the above time points, and you can forward it to your aunt with one click. If you want me to send on your behalf, you`d need to provide your aunt`s phone number and authorization, but current channels don`t support direct external messaging, so I`ll ensure no omissions through chat reminders.

Do you need to adjust the text message wording, or add one-click navigation link instructions for ride-hailing? If you have no other requirements, I`ll remind you and push text message templates at the above times.
\end{assistantbox}

\begin{userbox}
\textbf{[61]} No need to change, just remind and push as you said. Is there anything we haven`t arranged or missed now?
\end{userbox}

\begin{assistantbox}
\textbf{[62]} \textit{Tool Calls (1):} \\
\texttt{Tool 1: weather} \\
\textbf{Arguments:} \{"address": "Dalian Zhongshan District","date\_end": "2025-07-27","date\_start": "2025-07-27"\}
\end{assistantbox}

\begin{toolbox}
\textbf{[63]} \textit{Tool Response:} \\
\texttt{city: Dalian, weather: sunny, datetime: 2025-07-27, temperature: 26.0~34.0, humidity: 68.0}
\end{toolbox}

\begin{assistantbox}
\textbf{[64]} Overall arrangements are complete. Key items review:

Restaurant Reservation: Harbor Family Feast Restaurant, July 27th 12:00, 6 people, near accessible pathway, reservation number OIe6a7029150, paid.

Supply Delivery: JD Daojia | Yulong Premium

Walking cane (black adjustable) ×1

Adult diapers L size (10-piece pack) ×1

Expected departure 10:45, expected delivery around 12:02, delivered to restaurant front desk, order number OT464eb4905c, paid.

Aunt`s High-speed Rail: G901 (Beijing South 06:05 → Dalian North 10:47), first class ×1, order number OOba4980268b, paid.

Arrival Route: Dalian North Station → Harbor Family Feast Restaurant about 2.3km, recommend taxi, door-to-door more convenient.

Text Message Template \& Reminders:

10:35 Pre-arrival reminder (with taxi + restaurant address text message template for forwarding to aunt)

10:47 Arrival reminder (push text message template again)

Departure reminders: 05:15 departure, 05:55 ticket checking

Delivery reminder: 11:52 (about 10 minutes before expected delivery), if not received by 12:15 I`ll proactively assist with follow-up

Suggested Addition (optional but practical):

14:10 call taxi from restaurant to Dalian Port to ensure comfortable boarding. Do you need me to add a 14:10 taxi reminder?

Weather Alert: July 27th Zhongshan District sunny, 26-34°C, humidity about 68\%. Suggest preparing sunscreen and some drinking water, taking care of elderly and children`s hydration and sun protection.

If you agree, I`ll add the 14:10 departure reminder. Other than that, the itinerary is fully arranged. Do you need me to make any other adjustments?
\end{assistantbox}

\begin{userbox}
\textbf{[65]} Add the 14:10 taxi reminder, nothing else needed. Don`t forget to remind me when the time comes.
\end{userbox}

\begin{assistantbox}
\textbf{[66]} Added for you: July 27th 14:10 taxi reminder from Harbor Family Feast Restaurant to Dalian Port. I`ll push it in this chat on time and won`t forget.

Current Reminder List

05:15 Departure reminder for Beijing South Station

05:55 Pre-departure ticket checking reminder

10:35 Aunt`s pre-arrival reminder (with taxi + restaurant address text message template for forwarding)

10:47 Aunt`s arrival reminder (push text message template again)

11:52 Delivery expected arrival reminder about 10 minutes early

14:10 Taxi reminder from restaurant to Dalian Port

Everything is ready. If there are any temporary changes, I`ll synchronize immediately in this chat.

\#\#\#STOP\#\#\#
\end{assistantbox}

\end{document}